\documentclass{article}

\usepackage{microtype}
\usepackage{graphicx}
\usepackage{subfigure}
\usepackage{booktabs} 

\usepackage{hyperref}

\usepackage[accepted]{icml2023}

\usepackage{amsmath}
\usepackage{amssymb}
\usepackage{mathtools}
\usepackage{amsthm}

\usepackage[capitalize,noabbrev]{cleveref}

\theoremstyle{plain}

\theoremstyle{definition}

\theoremstyle{remark}

\newif\ifshowcomments 

\showcommentstrue  

\usepackage{url,wrapfig, subfigure}
\usepackage{caption}
\usepackage{graphicx}
\usepackage{xcolor}
\usepackage{amsmath}
\usepackage{setspace}
\usepackage{multirow}
\usepackage{wrapfig}
\usepackage{bm}
\usepackage{enumitem} 

\usepackage{xspace}
\usepackage{comment}

\usepackage{hyperref} 
\usepackage{algorithm} 
\usepackage{algorithmic} 
\usepackage{mathtools} 
\usepackage{amsthm, amssymb, amscd, amsfonts} 
\usepackage{thmtools}
\usepackage{thm-restate} 
\usepackage{wrapfig}
\usepackage{minitoc}  

\renewcommand{\thepage}{\arabic{page}}

\renewcommand{\algorithmicrequire}{\textbf{Input:}}
\renewcommand{\algorithmicensure}{\textbf{Output:}}

\DeclareMathOperator*{\argmin}{arg\,min\,}

\newcommand{\R}{\ensuremath{\mathbb{R}}}

\newcommand{\nvar}{D}
\newcommand{\ndim}{\nvar}

\newcommand{\x}{x}
\newcommand{\xvec}{\bm{\x}}

\newcommand{\y}{y}
\newcommand{\yvec}{\bm{\y}}

\newcommand{\z}{z}
\newcommand{\zvec}{\bm{\z}}

\newcommand{\E}{\mathbb{E}}

\providecommand{\sourceD}{P_{src}}
\providecommand{\targetD}{P_{tgt}}
\providecommand{\TIT}{T_{IT}}  
\providecommand{\Aset}{\mathcal{A}}

\providecommand{\IntSet}{\Omega}  
\providecommand{\HighDIntSet}{\Omega_{\text{high-dim}}}  
\providecommand{\kIntSet}{\IntSet^{(k)}}  
\providecommand{\tildeIntSet}{\tilde{\IntSet}}

\ifshowcomments
    \newcommand{\seannote}[1]{{\color{violet}Sean: #1}}
    \newcommand{\david}[1]{{\color{blue}David: #1}}
    
    \newcommand{\note}[1]{{\color{orange}{#1}}}  
    \newcommand{\askquestion}[1]{{\color{red}{#1}}}  
\else
    \newcommand{\seannote}[1]{}
    \newcommand{\david}[1]{}
    
    \newcommand{\note}[1]{}  
    \newcommand{\askquestion}[1]{}  
\fi

\providecommand{\ie}{i.e.\xspace}
\providecommand{\eg}{e.g.,\xspace}

\renewcommand{\th}{\textsuperscript{th}\xspace}

\providecommand{\argmin}{\mathop{\mathrm{argmin}}}

\providecommand{\st}{\text{s.t.}}

\providecommand{\R}{\ensuremath{\mathbb{R}}}

\providecommand{\E}{\mathbb{E}}

\providecommand{\Tpush}{T_\sharp}
\providecommand{\TOT}{T_{OT}}

\providecommand{\xvec}{\ensuremath{\mathbf{x}}}

\providecommand{\yvec}{\ensuremath{\mathbf{y}}}

\providecommand{\zvec}{\ensuremath{\mathbf{z}}}

\begin{document}
\twocolumn[
\icmltitle{Towards Explaining Distribution Shifts}



\icmlsetsymbol{equal}{*}

\begin{icmlauthorlist}
\icmlauthor{Sean Kulinski}{xxx}
\icmlauthor{David I. Inouye}{xxx}
\end{icmlauthorlist}

\icmlaffiliation{xxx}{Department of Electrical and Computer Engineering, Purdue University, West Lafayette, IN, USA}

\icmlcorrespondingauthor{David I. Inouye}{dinouye@purdue.edu}
\icmlcorrespondingauthor{Sean Kulinski}{skulinsk@purdue.edu}

\icmlkeywords{Machine Learning, ICML, Distribution Shift, Data-centric learning}

\vskip 0.3in
]

\printAffiliationsAndNotice{}  

\begin{abstract}
   A distribution shift can have fundamental consequences such as signaling a change in the operating environment or significantly reducing the accuracy of downstream models. Thus, understanding distribution shifts is critical for examining and hopefully mitigating the effect of such a shift. Most prior work focuses on merely detecting if a shift has occurred and assumes any detected shift can be understood and handled appropriately by a human operator. We hope to aid in these manual mitigation tasks by explaining the distribution shift using interpretable transportation maps from the original distribution to the shifted one. We derive our interpretable mappings from a relaxation of optimal transport, where the candidate mappings are restricted to a set of interpretable mappings. We then inspect multiple quintessential use-cases of distribution shift in real-world tabular, text, and image datasets to showcase how our explanatory mappings provide a better balance between detail and interpretability than baseline explanations by both visual inspection and our PercentExplained metric.
\end{abstract}

\vspace{-2 em}
\section{Introduction}
Most real-world environments are constantly changing, and in many situations, understanding how a specific operating environment has changed is crucial to making decisions respective to such a change.
Such a change might be due to a new data distribution seen in deployment which causes a machine-learning model to begin to fail.
Another example is a decrease in monthly sales data which could be due to a temporary supply chain issue in distributing a product or could mark a shift in consumer preferences.
When these changes are encountered, the burden is often placed on a human operator to investigate the shift and determine the appropriate reaction, if any, that needs to be taken.
In this work, our goal is to aid these operators in providing an explanation of such a change.

This ubiquitous phenomenon of having a difference between related distributions is known as distribution shift.
Much prior work focuses on \emph{detecting} distribution shifts; however, there is little prior work that looks into \emph{understanding} a detected distribution shift.
As it is usually solely up to an operator investigating a flagged distribution shift to decide what to do next, understanding the shift is critical for the operator to more efficiently mitigate any harmful effects of the distribution shift.
Due to the fact that there are no cohesive methods for understanding distribution shifts, as well as, the potential high complexity of distribution shifts (\eg \cite{koh2021wilds}), this important manual investigation task can be daunting.
The current de facto standard in analyzing a shift in tabular data is to look at how the mean of the original, \emph{source}, distribution shifted to the new, \emph{target}, distribution.
However, this simple explanation can miss crucial shift information due to being a coarse summary (\eg 
\autoref{fig:adult-income-results}) or, in high-dimensional regimes, can be uninterpretable.
Thus, there is a need for methods that can automatically provide detailed, yet interpretable, information about a detected shift which ultimately can lead to actionable insights about the shift.

\begin{figure*}[ht]
    \centering
    
    \includegraphics[width=\textwidth]{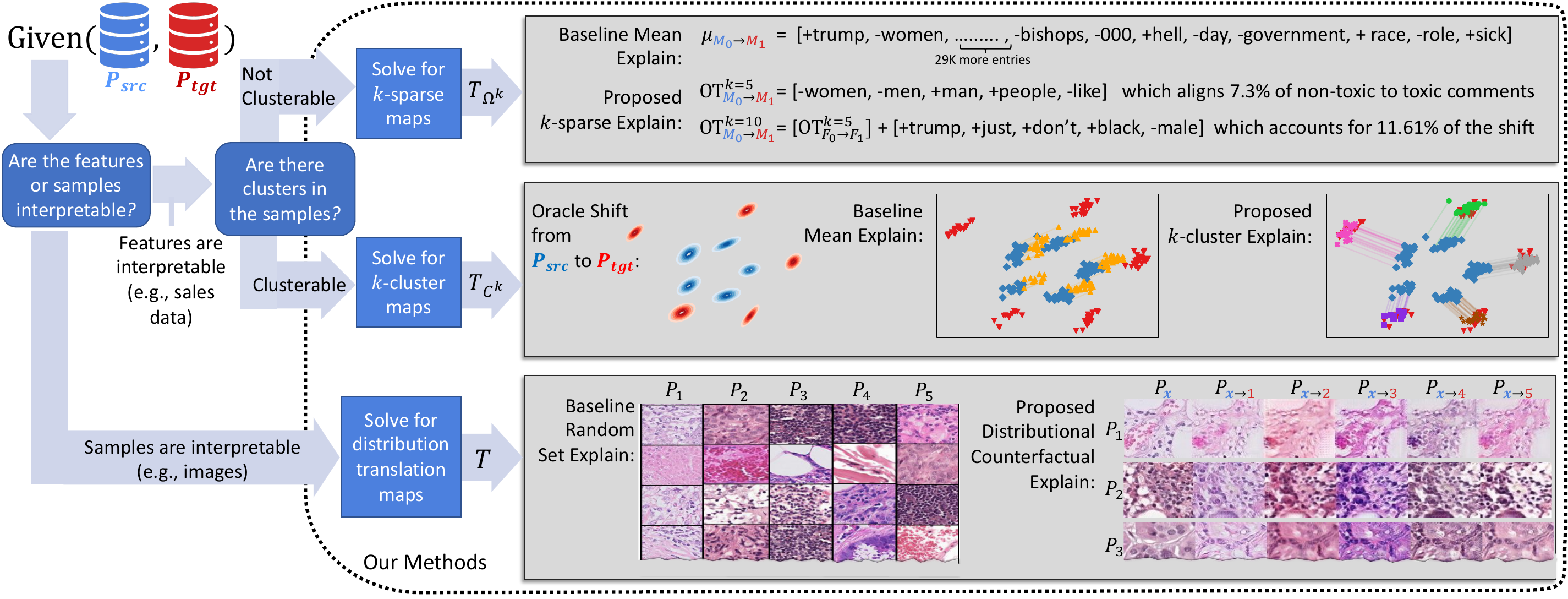}
    \vspace{-2 em}
    \caption{
    An overall look at our approach to explaining distribution shifts, where given a source $\sourceD$ and shifted $\targetD$ dataset pair, a user can choose to explain the distribution shift using $k$-sparse maps (which are best suited for high dimensional or feature-wise complex data), $k$-cluster maps (for tracking how heterogeneous groups change across the shift), or distribution translation maps (for data which has uninterpretable raw features such as images).
    For details on the results seen in the three boxes, please see experiments in \autoref{sec:experiments} and \autoref{sec:simulated-and-simple-experiments}.
    }
    \vspace{-1 em}
    \label{fig:main}
\end{figure*}

Therefore, we propose a novel framework for explaining distribution shifts, such as showing how features have changed or how groups within the distributions have shifted.
Since a distribution shift can be seen as a movement from a source distribution $\xvec \sim \sourceD$ to a target distribution $\yvec \sim \targetD$,
we define a distribution shift explanation as a transport map $T(\xvec)$ which maps a point in our source distribution onto a point in our target distribution.
For example, under this framework, the typical distribution shift explanation via mean shift can be written as $T(\xvec) = \xvec + (\mu_{\yvec} - \mu_{\xvec})$.
Intuitively, these transport maps can be thought of as a functional approximation of how the source distribution could have moved in a distribution space to become our target distribution.
However, without making assumptions on the type of shift, there exist many possible mappings that explain the shift (see \autoref{ssec:examples-of-infinite-mappings} for examples).
Thus, we leverage prior optimal transport work to define an ideal distribution shift explanation and develop practical algorithms for finding and validating such maps.

We summarize our contributions as follows:
\begin{itemize}[topsep=0pt,parsep=0pt,partopsep=0pt,leftmargin=*]
\item In \autoref{sec:explaining-distribution-shift}, we define intrinsically interpretable transport maps by constraining a relaxed form of the optimal transport problem to only search over a set of interpretable mappings and suggest possible interpretable sets. 
Also, we suggest methods for explaining image-based shifts such as distributional counterfactual examples.
\item In \autoref{sec:methods}, we develop practical methods and a PercentExplained metric for finding intrinsically interpretable mappings which allow us to adjust the interpretability of an explanation to fit the needs of a situation. 
\item In \autoref{sec:experiments}, we show empirical results on real-world tabular, text, and image-based datasets demonstrating how our explanations can aid an operator in understanding how a distribution has shifted.
\end{itemize}

\vspace{-0.5 em}
\section{Related Works}

\label{sec:related-works}

The characterization of the problem of distribution shift has been extensively studied \citep{quinonero2009dataset, storkey2009training, moreno2012unifying} via breaking down a joint distribution $P (\xvec, y)$ of features $\xvec$ and outputs $y$,  into conditional factorizations such as $P(y|\xvec) P(\xvec)$ or $P(\xvec|y) P(y)$. 
For covariate shift \citep{sugiyama2007covariate} the $P(\xvec)$ marginal differs from source to target, but the output conditional $P(y|\xvec)$ the same, while label shift (also known as prior probability shift) \citep{zhang2013domain, lipton2018detecting} is when the $P(y)$ marginals differ from source to target, but the full-feature conditional $P(\xvec | y)$ remains the same.
In this work, we refer to general problem distribution shift, \ie a shift in the joint distribution (with no distinction between $y$ and $\xvec$), and leave applications of explaining specific sub-genres of distribution shift to future work. 

As far as we are aware, this is the first work specifically tackling explaining distribution shifts, thus there are no accepted methods, baselines, or metrics for distribution shift explanations.
However, there are distinct works that can be applied to explain distribution shifts.
For example, one could use feature attribution methods \citep{saarela2021comparison, molnar2020interpretable} on a domain/distribution classifier (e.g., \citet{shanbhag2021unified} uses Shapley values \citep{shapley1997value} to explain how changing input feature distributions affect a classifier's behavior), or once could find samples which are most illustrative of the differences between distributions \citep{brockmeier2021identifying}.
Additionally, one could use counterfactual generation methods \citep{karras2019style, sauer2021counterfactual, pawelczyk2020learning} and apply them for ``distributional counterfactuals'' which would show what a sample from $\targetD$ would have looked like if it instead came from $\sourceD$ (e.g., \citet{pawelczyk2020learning} uses a classifier-guided VAE to generate class counterfactuals on tabular data).
We explore this distributional counterfactual explanation approach in \autoref{ssec:counterfactual-examples}.

A sister field is that of detecting distribution shifts.
This is commonly done using methods such as statistical hypothesis testing of the input features \citep{nelson2003applied, rabanser2018failing, quinonero2009dataset}, training a domain classifier to test between source and non-source domain samples \citep{lipton2018detecting}, etc.
In \citet{kulinski2020feature, budhathoki2021why}, the authors attempt to provide more information beyond the binary ``has a shift occurred?'' via localizing a shift to a subset of features or causal mechanisms.
\citet{kulinski2020feature} does this by introducing the notion of Feature Shift, which first detects if a shift has occurred and if so, localizes that shift to a specific subset of features that have shifted from source to target.
In \citet{budhathoki2021why}, the authors take a causal approach via individually factoring the source and target distributions into a product of their causal mechanisms (\ie a variable conditioned on its parents) using a shared causal graph, which is assumed to be known \emph{a priori}.
Then, the authors ``replace'' a subset of causal mechanisms from $\sourceD$ with $\targetD$, and measure divergence from $\sourceD$ (i.e. measuring how much the subset change affects the source distribution).
Both of these methods are still focused on detecting distribution shifts (via identifying shifted causal mechanisms or feature-level shifts), unlike explanatory mappings which help explain \emph{how} the data has shifted.

\vspace{-0.5 em}
\section{Explaining Shifts via Transport Maps}
\label{sec:explaining-distribution-shift}
\vspace{-0.25 em}
The underlying assumption of distribution shift is that there exists a relationship between the source and target distributions.
From a distributional standpoint, we can view distribution shift as a \emph{movement}, or transportation, of samples from the source distribution $\sourceD$ to the target distribution $\targetD$.
Thus, we can capture this relationship between the distributions via a transport map $T$ from the source distribution to the target, i.e., if $\xvec \sim \sourceD$, then $T(\xvec) \sim \targetD$.
If this mapping is understandable to an operator investigating a distribution shift, this can serve as an explanation to the operator on what changed between the environments; thus allowing for more effective reactions to the shift.
Therefore, in this work, we define a distribution shift explanation as: \emph{finding an interpretable transport map $T$ which approximately maps a source distribution $\sourceD$ onto a target distribution $\targetD$ such that $\Tpush \sourceD \approx \targetD$}.
Similar to ML model interpretability \cite{molnar2020interpretable}, an interpretable map can either be one that is intrinsically interpretable (\autoref{subsec:intrinsically-interpretable-transportation-maps}) or a mapping that is explained via post-hoc methods such as sets of input-output pairs (\autoref{ssec:counterfactual-examples}).

\vspace{-0.5 em}
\subsection{Intrinsically Interpretable Transportation Maps}
\vspace{-0.5 em}
\label{subsec:intrinsically-interpretable-transportation-maps}
To find such a mapping between distributions, it is natural to look to Optimal Transport (OT) and its extensive prior work in this field \cite{cuturi2013sinkhorn, arjovsky2017wasserstein, torres2021survey, peyre2019Computational}.
An OT mapping given a transportation cost function $c$ is a method of optimally moving points from one distribution to align with another distribution and is defined as:
\vspace{-0.5 em}
\begin{equation*}
\label{eq:OTContinuous}
    \TOT \coloneqq  \argmin_T \E_{\sourceD} \left[ c(\xvec, T(\xvec)) \right] ~ \st \,\, \Tpush \sourceD = \targetD
\end{equation*}
where $\Tpush \sourceD$ is the pushforward operator that can be viewed as applying  $T$ to all points in $\sourceD$, and $\Tpush \sourceD = \targetD$ is the marginal constraint, which means the pushforward distribution must match the target distribution.
OT is a natural starting point for shift explanations as it allows for a rich geometric structure on the space of distributions, and by finding a mapping that minimizes a transport cost we are forcing our mapping to retain as much information about the original $\xvec$ samples when aligning $\sourceD$ and $\targetD$.
For more details about OT, please see \cite{villani2009optimal, peyre2019Computational}. 

However, since OT considers all possible mappings which satisfy the marginal constraint, this means the resulting $\TOT$ can be arbitrarily complex and thus possibly uninterpretable as a shift explanation.
We can alleviate this by restricting the candidate transport maps to belong to a set of user-defined interpretable mappings $\IntSet$.
However, this problem can be infeasible if $\IntSet$ does not contain a mapping that satisfies the marginal alignment constraint.
Therefore, we can use Lagrangian relaxation to relax the marginal constraint, giving us an \emph{Interpretable Transport} mapping $\TIT$:
\vspace{-0.3 em}
\begin{equation}
\label{eq:Interpretable-Transport}
        \TIT \coloneqq \argmin_{T \in \IntSet} ~ \E_{\sourceD} \left [ c ( \xvec, T(\xvec) ) \right] + \lambda ~ \phi(P_{T(\xvec)}, \targetD)
\end{equation}

where $\phi( \cdot, \cdot )$ is a distribution divergence function (e.g., KL or Wasserstein).
In this paper, we will assume $c$ is the squared Euclidean cost and $\phi( \cdot, \cdot )$ is the squared Wasserstein-2 metric, unless stated otherwise.
Due to the heavily complex and context-specific nature of distribution shift, it is likely that $\IntSet$ would be initialized based on context.
However, we suggest two \emph{general} methods in the next section as a starting point and hope that future work can build upon this framework for specific contexts.

\vspace{-0.5 em}
\subsection{Intrinsically Interpretable Transport Sets}
\label{sec:interpretable-transport-sets}
\vspace{-0.5 em}

The current common practice for explaining distribution shifts is comparing the means of the source and the target distributions.
The mean shift explanation can be generalized as $\Omega_{\text{vector}} = \{ T : T(\xvec) = \xvec + \delta \}$ where $\delta$ is a constant vector and mean shift being the specific case where $\delta$ is the difference of the source and target means.
By letting $\delta$ be a function of $\xvec$, which further generalizes the notion of mean shift by allowing each point to move a variable amount per dimension, we arrive at a transport set that includes any possible mapping $T: \R^D \to \R^D$.
However, even a simple transport set like $\Omega_{\text{vector}}$ can yield uninterpretable mappings in high dimensional regimes (e.g., a shift vector of over 100 dimensions).
To combat this, we can constrain the complexity of a mapping by forcing it to only move points along a specified number of dimensions, which we call \emph{$k$-Sparse Transport}.
 
\textbf{$k$-Sparse Transport:}
For a given class of transport maps, $\Omega$ and a given $k \in  \{1, ..., D\}$, we can find a subset $\Omega^{(k)}_{sparse}$ which is the set of transport maps from $\Omega$ which only transport points along $k$ dimensions or less.
Formally, we define an active set $\Aset$ to be the set of dimensions along which a given $T$ moves points: $\Aset(T) \triangleq  \{j \in \{1,\dots,D\}: \exists \xvec, T(\xvec)_j - x_j \neq 0 \}$. Then, we define $\Omega^{(k)}_{sparse} = \{T \in \Omega: |\Aset(T)| \leq k \}$.

$k$-sparse transport is most useful in situations where a distribution shift has happened along a subset of dimensions, such as explaining a shift where some sensors in a network are picking up a change in an environment.
However, in cases where points shift in different directions based on their original value, e.g. when investigating how a heterogeneous population responded to an advertising campaign, $k$-sparse transport is not ideal.
Thus, we provide a shift explanation that breaks the source and target distributions into $k$ sub-populations and provides a vector-based shift explanation per sub-population, which we call $k$-\emph{Cluster Transport}.

\textbf{$k$-Cluster Transport:} Given a $k \in \{1, \dots, D\}$ we define $k$-cluster transport to be a mapping which moves each point $\xvec$ by constant vector which is specific to $\xvec$'s cluster.
More formally, we define a labeling function $\sigma(\xvec; M) \triangleq \argmin_j \|\bm{m}_j - \xvec\|_2$, which returns the index of the column in $M$ (\ie the label of the cluster) which $\xvec$ is closest to.
With this, we define $\Omega_{\text{cluster}}^{(k)} =$ $\left \{T : T(\xvec) = \xvec + \delta_{\sigma(\xvec;M)}, M\in \R^{D\times k}, \Delta \in \R^{D \times k} \right \}$, where $\delta_j$ is the $j$\th column of $\Delta$. 

Since measuring the exact interpretability of a mapping is heavily context-dependent, we can instead use $k$ in the above transport maps to define a partial ordering of interpretability of mappings \emph{within} a class of transport maps.
Let $k_1$ and $k_2$ be the size of the active sets for $k$-sparse maps (or the number of clusters for $k$-cluster maps) of $T_1$ and $T_2$ respectively.
If $k_1 \leq k_2$, then $\text{Inter}(T_1) \geq \text{Inter}(T_2)$, where $\text{Inter}(T)$ is the interpretability of shift explanation $T$.
For example, we claim the interpretability of a $T_1 \in \Omega_{sparse}^{(k=10)}$ is greater than (or possibly equal to) the interpretability of a $T_2 \in \Omega_{sparse}^{(k=100)}$ since a shift explanation in $\Omega$ which moves points along only 10 dimensions is more interpretable than a similar mapping which moves points along 100 dimensions.
A similar result can be shown for $k$-cluster transport since an explanation of how 5 clusters moved under a shift is less complicated than an explanation of how 10 clusters moved.
The above method allows us to define a partial ordering on interpretability without having to determine the absolute value of interpretability of an individual explanation $T$, as this requires expensive context-specific human evaluations, which are out of scope for this paper.

\vspace{-0.5 em}
\subsection{Intrinsically Interpretable Maps For Images}
\vspace{-0.5 em}
\label{ssec:interpretable-mappings-for-images}
To find interpretable transport mappings for images, we could first project $\sourceD$ and $\targetD$ onto a low-dimensional \emph{interpretable} latent space (\eg a space which has disentangled and semantically meaningful dimensions) and then apply the methods above in this \emph{latent} space.
Concretely, let us denote the \mbox{(pseudo-)invertible} encoder as $g: \R^D \rightarrow \R^{D'}$ where $D' < D$ (e.g., an autoencoder).
Given this encoder, we define our set of high dimensional interpretable transport maps: $\HighDIntSet \coloneqq \left\{ T : T= g^{-1}\left( \tilde{T}\left( g(\xvec) \right) \right), \tilde{T} \in \IntSet, g \in \mathcal{I} \right\}$ where $\IntSet$ the set of interpretable mappings (\eg $k$-sparse mappings) and $\mathcal{I}$ is the set of \mbox{(pseudo-)invertible} functions with an interpretable (\ie semantically meaningful) latent space. 
Finally, given an interpretable $g \in \mathcal{I}$, this gives us \emph{High-dimensional Interpretable Transport}: $T_{HIT}$.

As seen in the Stanford Wilds dataset \cite{koh2021wilds}, which contains benchmark examples of real-world image-based distribution shifts, image-based shifts can be immensely complex.
In order to provide an adequate intrinsically interpretable mapping explanation of a distribution shift in high dimensional data (\eg images), multiple new advancements must first be met (\eg finding a disentangled latent space with semantically meaningful dimensions, approximating high dimensional empirical optimal transport maps, etc.), which are out of scope of this paper.
We further explore details about $T_{HIT}$, its variants, and the results of using $T_{HIT}$ to explain Colorized-MNIST in \autoref{sec:appendix-explaining-shifts-in-images}, and we hope future work could build upon this framework.

\vspace{-0.5 em}
\subsection{Post-Hoc Explanations of Image-Based Mappings via Counterfactual Examples}
\label{ssec:counterfactual-examples}
\vspace{-0.5 em}
As mentioned above, in some cases, solving for an interpretable latent space can be too difficult or costly, and thus a shift cannot be expressed by an interpretable mapping function.
However, if the samples themselves are easy to interpret (e.g., images), we can still explain a transport mapping by visualizing translated samples.
Specifically, we can remove the interpretable constraint on the mapping itself and leverage methods from the unpaired Image-to-Image translation (I2I) literature to translate between the source and target domain while preserving the content.
For a comprehensive summary of the recent I2I works and methods, please see \cite{pang2021image}.

Once an I2I mapping is found, to serve as an explanation, we can provide an operator with a set of counterfactual pairs $\left\{ ( \xvec, T(\xvec) ) : \xvec \sim \sourceD, T(\xvec) \sim \targetD \right\}$.
Then, by determining what commonly stays invariant and what commonly changes across the set of counterfactual pairs, this can serve as an explanation of how the source distribution shifted to the target distribution.
While more broadly applicable, this approach could put a higher load on the operator than an intrinsically interpretable mapping approach.

\section{Practical Methods for Finding and Validating Shift Explanations}
\label{sec:methods}

In this section, we discuss practical methods for finding these maps via empirical OT (Sec.~\ref{sec:emperical-interpretable-transport}, \ref{ssec:sparse-explanation-maps}, and \ref{ssect:cluster-based-shift}) and introduce a PercentExplained metric which can assist the operator in selecting the hyperparameter $k$ in $k$-sparse and $k$-cluster transport (Sec.~\ref{sec:interpretability-as-a-hyperparameter}).

\vspace{-0.5 em}
\subsection{Empirical Interpretable Transport Upper Bound}
\vspace{-0.5 em}
\label{sec:emperical-interpretable-transport}
As the divergence term in our interpretable transport objective (\autoref{eq:Interpretable-Transport}) can be computationally-expensive to optimize in practice, we propose to optimize the following simplification, which simply computes the difference between the map and the sample-based OT solution $\TOT$ (which can be computed efficiently for samples or approximated via the Sinkhorn algorithm \citep{cuturi2013sinkhorn}):
\begin{equation}
\label{eq:emperical-interpretable-transport}
\argmin_{T \in \IntSet} \! \frac{1}{N} \!\sum_{i=1}^N c \big( \xvec^{(i)}, T( \xvec^{(i)} ) \big) + 
\lambda d \big( T (\xvec^{(i)}), \TOT (\xvec^{(i)}) \big)
\end{equation}
where $d$ is the squared $\ell_2$ function.
Notably, the divergence value in \autoref{eq:Interpretable-Transport} is replaced with the average over a sample-specific distance between $T(\xvec)$ and the optimal transport mapping $\TOT(\xvec)$.
This is computationally attractive as the optimal transport solution only needs to be calculated once, rather than calculating the Wasserstein distance once per iteration as would be required if directly optimizing the Interpretable Transport problem.
Additionally, we prove in \autoref{ssec:emperical-IT-upper-bound} that the second term in \autoref{eq:emperical-interpretable-transport} is an upper bound when the divergence is the squared Wasserstein distance, i.e., when $\phi=W_2^2$.

\vspace{-0.5 em}
\subsection{Finding $k$-Sparse Maps} 
\label{ssec:sparse-explanation-maps}
\vspace{-0.5 em}
The $k$-sparse algorithm can be broken down into two steps.
First, given $k$, we estimate the active set $\Aset$ by simply taking the $k$ dimensions with the largest difference of means between two distributions.
This is a simple approach that avoids optimization over an exponential number of possible subsets for $\Aset$ and can be optimal for some cases, as explained below.
Second, given the active set $\Aset$, we need to estimate the map.
While estimating $k$-sparse solutions to the original interpretable transport problem (\autoref{eq:Interpretable-Transport}) is challenging, we prove that the solution with optimal alignment to the upper bound above (\autoref{eq:emperical-interpretable-transport}) can be computed in closed-form for two special cases.
If the optimization set is restricted to only shifting the mean, i.e., $\kIntSet = \Omega_{vector}^{(k)}$, then the solution with optimal alignment is:

\begin{align}
    \forall j, [T(\xvec)]_j = \left \{ 
    \begin{array}{ll}
    x_j + (\mu_{j}^{\text{tgt}} - \mu_{j}^{\text{src}}), & \text{if} \,\, j \in \Aset \\
    x_j, & \text{if} \,\, j \not\in \Aset \\
    \end{array}
    \right. \,,
\end{align} 
where $\mu^{\text{src}}$ and $\mu^{\text{tgt}}$ are the mean of the source and target distributions respectively.
Similarly, if $\kIntSet$ is unconstrained except for sparsity, then the solution with optimal alignment is simply:
\begin{align}
    \forall j, [T(\xvec)]_j = \left \{ 
    \begin{array}{ll}
    [\TOT(\xvec)]_j, & \text{if} \,\, j \in \Aset \\
    x_j, & \text{if} \,\, j \not\in \Aset \\
    \end{array}
    \right . \,,
\end{align} 
where $[\TOT(\xvec)]_j$ is the $j$-th coordinate of the sample-based OT solution.
The proofs of alignment optimality w.r.t. the divergence upper bound in \autoref{eq:emperical-interpretable-transport} are based on decomposability of the squared Euclidean distance and can be found in \autoref{sec:appendix-proofs}.
The final algorithm for both sparse maps can be found in \autoref{alg:k-sparse-alg}.

\begin{algorithm}[!h]
\caption{Finding $k$-Sparse Maps}
\label{alg:k-sparse-alg}
\begin{algorithmic}
\STATE \textbf{Input:} Domain datasets $X \in \mathbb{R}^{N \times \ndim}$ and $Y^{N \times \ndim}$ with $N$ samples of dimensionality $\ndim$ each, the desired sparsity $k$, and interpretable set type, i.e., $\IntSet$.
\vspace{0.5em}
\STATE \emph{// Select active set based on means}
\STATE $\mu^{\text{diff}} \gets \mu^{\text{tgt}}-\mu^{\text{src}} = \frac{1}{N} \sum_{i=1}^{N} Y_i - \frac{1}{N} \sum_{i=1}^{N} X_i$
\STATE $\Aset \gets \textnormal{TopKIndices}(\text{abs}(\mu^{\text{diff}}), k)$
\STATE \emph{// Create dimension-wise maps based on active set}
\IF{$\IntSet = \Omega_{vector}$}
    \STATE $\forall j, [T(\xvec)]_j = \left \{ 
    \begin{array}{ll}
    x_j + \mu^{\text{diff}}_j, & \text{if} \,\, j \in \Aset \\
    x_j, & \text{if} \,\, j \not\in \Aset \\
    \end{array}
    \right.$
\ELSE 
    \STATE $\TOT(\cdot) \gets \textnormal{OptimalTransportAlg}(X, Y)$
    \STATE $\forall j, [T(\xvec)]_j = \left \{ 
    \begin{array}{ll}
    [\TOT(\xvec)]_j, & \text{if} \,\, j \in \Aset \\
    x_j, & \text{if} \,\, j \not\in \Aset \\
    \end{array}
    \right.$
\ENDIF
\STATE \textbf{Output:} $T(\cdot)$
\end{algorithmic}
\end{algorithm}

\subsection{Finding $k$-Cluster Maps} 
\label{ssect:cluster-based-shift}

Similar to $k$-sparse maps, we split this algorithm into two parts: (1) estimate pairs of source and target clusters and then (2) compute mean shift for each pair of clusters.
For the first step, na\"ively one might expect that independent clustering on each domain dataset followed by post-hoc pairing of these clusters may be sufficient.
However, this could yield very poor clustering pairs that are significantly mismatched because the domain-specific clustering may not be optimal in terms of the alignment objective.
For example, the source domain may have one large and one small cluster and the target domain could have equal-sized clusters.
Therefore, it is important to cluster the source and domain samples jointly.
To estimate paired (i.e., dependent) clusterings of the source and target domain samples, we first find the OT mapping from source to target.
We then cluster an paired dataset formed by concatenating each source sample with its OT mapped sample (which actually corresponds to one of the target samples).
The clustering on these paired samples gives paired cluster centroids for the source and target, denoted $\mu_\ell^{\text{src}}$ and $\mu_\ell^{\text{tgt}}$ respectively, which we use to construct a cluster-specific mean shift map defined as:
\begin{align}
    T(\xvec) = \xvec + (\mu^{\text{tgt}}_{\sigma(\xvec)} - \mu^{\text{src}}_{\sigma(\xvec)})
\end{align}
where $\sigma(\xvec) = \argmin_\ell \|\xvec - \mu_\ell^{\text{src}}\|_2^2$ is the cluster label function.
This map applies a simple shift to every source domain cluster to map to the target domain.
\autoref{alg:cluster-alg} shows pseudo-code for both steps in our $k$-cluster method.

\newlength{\textfloatsepsave}
\setlength{\textfloatsepsave}{\textfloatsep}
\setlength{\textfloatsep}{8pt}
\begin{algorithm}[!h]
\caption{Solving for $k$-Cluster Mappings}
\label{alg:cluster-alg}
\begin{algorithmic}
\STATE \textbf{Input:} Domain datasets $X \in \mathbb{R}^{N \times \ndim}$ and $Y^{N \times \ndim}$ with $N$ samples of dimensionality $\ndim$ each and the desired number of clusters $k$.
\vspace{0.5em}
\STATE \textit{// Compute sample-based optimal transport map}
\STATE $\TOT(\cdot) \gets \text{OptimalTransportAlg}(X, Y)$
\STATE \emph{// Compute paired clustering}
\STATE $Z \gets [X, \TOT(X)] \in \mathbb{R}^{N \times 2\ndim}$
\STATE $[\mu_1,\cdots,\mu_k]^T \gets \text{KMeansClust}(Z, k) \in \mathbb{R}^{k \times 2\ndim}$
\STATE \emph{// Extract paired source and target centroids}
\STATE $\forall \ell \in \{1,\cdots,k\}, \mu_\ell^{\text{src}} = [\mu_{\ell,1},\cdots, \mu_{\ell,\ndim}]^T \in \mathbb{R}^\ndim$
\STATE $\forall \ell \in \{1,\cdots,k\}, \mu_\ell^{\text{tgt}} = [\mu_{\ell,\ndim + 1},\cdots, \mu_{\ell,2\ndim}]^T \in \mathbb{R}^\ndim$
\STATE \emph{// Setup final cluster-based map}
\STATE $\sigma(\xvec) = \argmin_\ell \|\xvec - \mu_\ell^{\text{src}}\|_2^2$ \hspace{0.5em} \emph{// Clust. label func}
\STATE $T(\xvec) = \xvec + (\mu^{\text{tgt}}_{\sigma(\xvec)} - \mu^{\text{src}}_{\sigma(\xvec)})$
\STATE \textbf{Output:} $T(\cdot)$
\end{algorithmic}
\end{algorithm}

\subsection{Interpretability as a Hyperparameter} 
\label{sec:interpretability-as-a-hyperparameter}
We now discuss how the $k$ hyperparameter in $k$-sparse and $k$-cluster maps can be adjusted to allow a user to automatically change the level of interpretability of a shift explanation as desired.
While an optimal shift explanation could be achieved by solving \autoref{eq:Interpretable-Transport}, directly defining the set $\Omega$, which should contain both interpretable yet sufficiently expressive maps, can be a difficult task.
Thus, we can instead set $\Omega$ to be a super-class, such as $\Omega_{vector}$ given in \autoref{sec:interpretable-transport-sets} and adjust $k$ until a $\kIntSet$ is found which matches the needs of the situation.
This allows a human operator to request a mapping with better alignment by increasing $k$, which correspondingly will decrease the mapping's interpretability, or request a more interpretable mapping by decreasing the complexity (\ie decreasing $k$) which will decrease the alignment.

To assist an operator in determining if the interpretability hyperparameter should be adjusted, we introduce a \emph{PercentExplained} metric, which we define to be: 
\begin{equation}
    \textnormal{PE}(\sourceD, \targetD, T) \coloneqq \frac{W_2^2(\sourceD, \targetD) - W_2^2(\Tpush \sourceD, \targetD)}{W_2^2(\sourceD, \targetD)}
\end{equation}
where $W_2^2(\cdot, \cdot)$ is the squared Wasserstein-2 distance between two distributions and PE is shorthand for PercentExplained.
By rearranging terms we get 
$1 - \frac{W_2^2 (\Tpush \sourceD, \targetD)}{W_2^2 ( \sourceD, \targetD)}$,
which shows this metric's correspondence to the statistics coefficient of determination $R^2$, where $W_2^2 (\Tpush \sourceD, \targetD)$ is analogous to the residual sum of squares and $W_2^2 ( \sourceD, \targetD)$ is similar to the total sum of squares.
This gives an approximation of how much a current shift explanation $T$ accurately maps onto a target distribution.
This can be seen as a normalization of a mapping's fidelity with the extremes being $\Tpush \sourceD = \targetD$, which fully captures a shift, and $T=\text{Id}$, which does not move the points at all.
When provided this metric along with a shift explanation, an operator can decide whether to accept the explanation (\eg the PercentExplained is sufficient and $T$ is still interpretable) or reject the explanation and adjust $k$. 

\setlength{\textfloatsep}{\textfloatsepsave}  
\begin{figure*}[!h]
    \centering
    \includegraphics[width=0.95\textwidth]{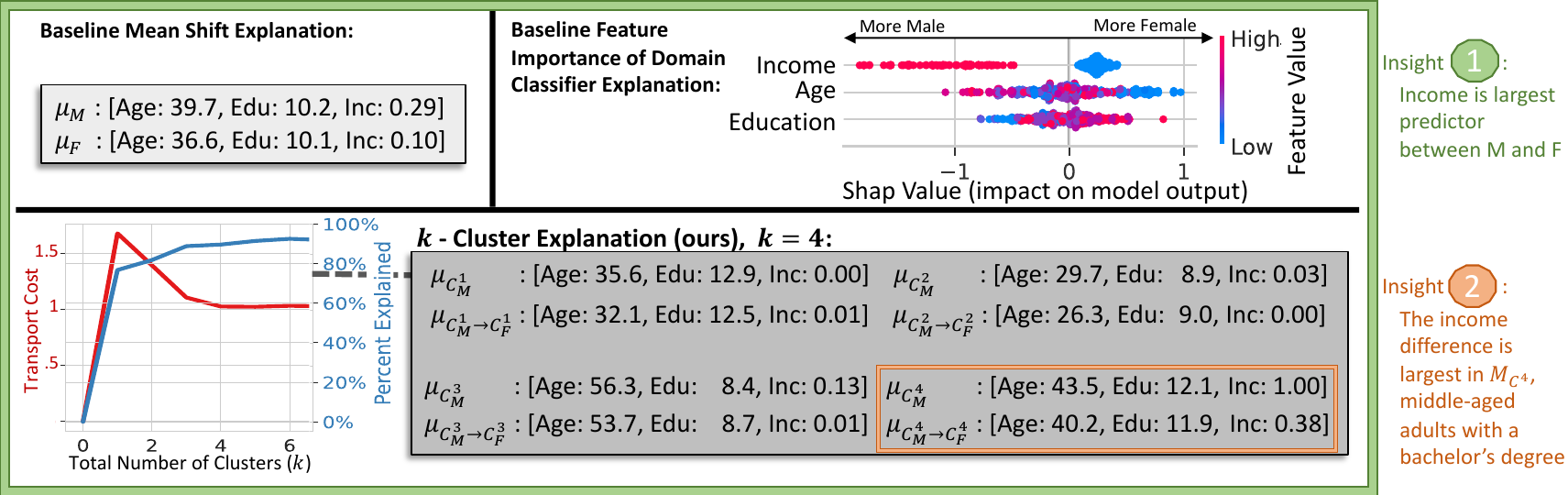}
    \vspace{-0.5 em}    
    \caption{Using $k$-cluster transport (bottom) to explain the shift from the male population to the female population of the Adult Income dataset allows us to capture how heterogeneous groups within the dataset moved.
    For example, while all three methods show that income is indeed the largest difference between $M$ and $F$ for this dataset ({\color{green} \textbf{insight 1}}), only the $k$-cluster-based explanation reveals {\color{orange} \textbf{insight 2}}, that the income disparity is most prevalent between middle-aged males and females with a bachelor's degree (edu=$12$) seen in $C^4$.
    }
    \vspace{-1 em}
    \label{fig:adult-income-results}
\end{figure*}

\vspace{-0.5 em}
\section{Experiments}
\label{sec:experiments}
\vspace{-0.5 em}

 In this section, we study the performance of our methods when applied to real-world data.\footnote{Code to recreate the experiments can be found at \href{https://github.com/inouye-lab/explaining-distribution-shifts}{https://github.com/inouye-lab/explaining-distribution-shifts}.}
 For gaining intuition on different explanation techniques, we point the reader to \autoref{sec:simulated-and-simple-experiments} where we present experiments on simple simulated shifts.
We first present results using $k$-cluster transport to explain the difference between different groups of the male population and groups of the female population in the U.S. Census ``Adult Income'' dataset \cite{kohavi1996uci}.
We then use $k$-sparse transport to explain shifts between toxic and non-toxic comments across splits from the Stanford WILDS distribution shift benchmark \cite{koh2021wilds} version of the ``CivilComments'' Dataset \cite{borkan2019nuanced}.
Finally, we use distributional counterfactuals to explain the high-dimensional shifts between histopathology images from different hospitals as seen in the WILDS Camelyon17 dataset \cite{bandi2018detection}.

\vspace{-1 em}
\paragraph{Adult Income Dataset}
This dataset originally comes from the United States 1994 Census database and is commonly used to predict whether a person's annual income exceeds \$50k using 14 demographic features.
Similar to \cite{budhathoki2021why}, we consider a subset of non-redundant features: age, years of education (where 12+ is beyond high school), and income (which is encoded as $1$ if the person's annual income is greater than \$50k and $0$ if it is below).
We then split this dataset along the sex dimension, and define our source distribution as the male population and the target as the female population.
In order to find the set of paired clusters, we first standardize a copy of the data to have zero mean and unit variance across all features, where the $\mu$ and $\sigma$ used for the standardization are found via the feature-wise mean and standard deviation of the source distribution and perform clustering in the standardized joint space using the method described in Section \ref{ssect:cluster-based-shift}.
The $k$ clustering labels are then used to label points to clusters in the original (unstandardized) data space.

Suppose our role is a researcher seeking to implement a social program targeting gender inequalities.
We could compare the means of the male/female distributions, which shows on average a 20\% lower chance of having an annual income above \$50k when moving from the male population to the female population.
Additionally, we could train a classifier to predict between male/female data points and use a feature importance measurement tool like Shapley values \cite{lundberg2017unified} to determine that income is a main differing feature ({\color{green} \textbf{insight 1}} from \autoref{fig:adult-income-results}).
However, suppose we want to dig deeper.
We could instead use $k$-cluster transport to see how heterogeneous subgroups shifted across a range of clusters, as seen in \autoref{fig:adult-income-results}.
If we accept the explanation at $k=4$ (since beyond this, the marginal advantage of adding an additional cluster is minimal in terms of both transport cost and PercentExplained), we can now make {\color{orange} \textbf{insight 2}}.
Here, $\mu_{C_M^4 \to C_F^4}$ shows the income difference is significantly larger between middle-aged males/females with a bachelor's degree (a decrease from nearly 100\% high-income likelihood to only a 38\% chance).
While {\color{green} \textbf{insight 1}} validates the need for our social program, {\color{orange} \textbf{insight 2}} (which is hidden in both the mean-shift and distribution classifier explanations) provides a significantly narrower scope for us to focus our efforts in, thus allowing for swifter action.

\vspace{-1 em}
\paragraph{Civil Comments Dataset}
Here we present results using $k$-sparse shifts to explain the difference between three splits of the CivilComments dataset \cite{borkan2019nuanced} from the WILDS datasets \cite{koh2021wilds}. 
This dataset consists of comments scraped from the internet where each comment is paired with a binary toxicity label and demographic information pertaining to the content of the comment.
If we were an operator trying to see how the comments and their toxicity change across targeted demographics, we could create three splits: \{F, M\}, \{F$_0$, F$_1$\}, and \{M$_0$, M$_1$\}, where F represents all female comments, M are all male comments, and F$\textsubscript{0}$, F$\textsubscript{1}$ are nontoxic, toxic female comments, respectively (and likewise for males). 
 We can explain these three splits using vanilla mean shift, a $k$-sparse mean shift ($k$-$\mu$), and $k$-sparse OT ($k$-OT) shift explanations, as seen in \autoref{tab: civil-comments} which shows results for the unigrams which the maximize the alignment between the unigram distributions created for each split.
 The baseline vanilla mean-shift explanation yields all 30K features at once (with no guide for truncating), while the $k$-sparse shifts provide explanations up to $k$ words as well as a corresponding PercentExplained to aid in determining if additional words should be added to the explanation.
Note that for $k$-$\mu$ explanations, when transporting a word, that word is added equally to all comments in $\sourceD$, while since $k$-OT allows for each comment to be shifted optimally (via conditioning on the other words in each comment), thus $k$-OT can explain significantly more of the shift by transporting the same word (which can highlight words that have complex dependencies such as ``don't'').

\begin{table*}[ht!]
\caption{A baseline vanilla mean shift explanation, $k$-sparse mean shift explanation, ($k$-$\mu$-Ex), and $k$-sparse OT explanations ($k$-OT-Ex) for the three splits from CivilComments (to save space the baseline is only used for F$\rightarrow$M).
Each cell represents adding/subtracting a unigram from $\sourceD$ to align it with the comment distribution of $\targetD$ and the respective PercentExplained (excluding the baseline method).
For example, in $k$-$\mu$-Ex(F\textsubscript{0}$\rightarrow$F\textsubscript{1}), adding ``stupid'' aligns the non-toxic female comments to the toxic female comments and cumulatively explains $0.2\%$ of the shift.
}
\vspace{-0.75 em}
\label{tab: civil-comments}
\includegraphics[width=\textwidth]{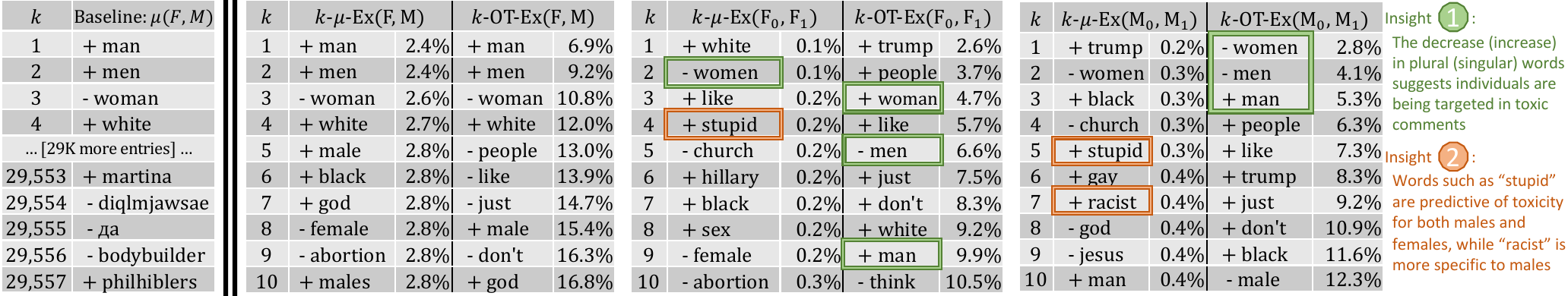}
\vspace{-2 em}
\end{table*}

It is clear that performing a vanilla mean shift explanation on the unigram data between splits is unwise due to the high dimensionality of the data and it is unclear when to truncate such an explanation. 
However, in our approach, by iteratively reporting the shifted unigrams along with the cumulative PercentExplained, a practitioner can better understand the impact each additional word has on the shift explanation.
For example, it makes sense that adding ``man'', ``men'', and subtracting ``woman'' were the three unigrams that best aligned the female and male comment distributions and could account for as much as 10\% of the shift.

With the $k$-sparse explanation, {\color{green} \textbf{insight 1}} suggests that a content moderator is more likely to encounter toxic comments that target individuals rather than groups, and thus a moderator could train a classifier to predict if the object of a comment is an individual or group of people. 
This quality, which is not obvious at first glance (and especially not from the simple mean shift explanation), may enable more explainable moderation as an explanation about why a comment was removed could state that a comment is targeting an individual as one feature.
Additionally, {\color{orange} \textbf{insight 2}} shows that if the moderator's goal is to be equitable across groups, they may want to target words that have roughly equal impact across groups like ``stupid''. 
If on the other hand, the moderator's goal is merely to detect any toxicity regardless of group, this may signal that they should provide group variables such as ``gay'' or ``racist'' (which are more predictive of toxicity towards males) to the model as it is an important signal.

\begin{figure}[!h]
    \centering
    \includegraphics[width=\columnwidth]{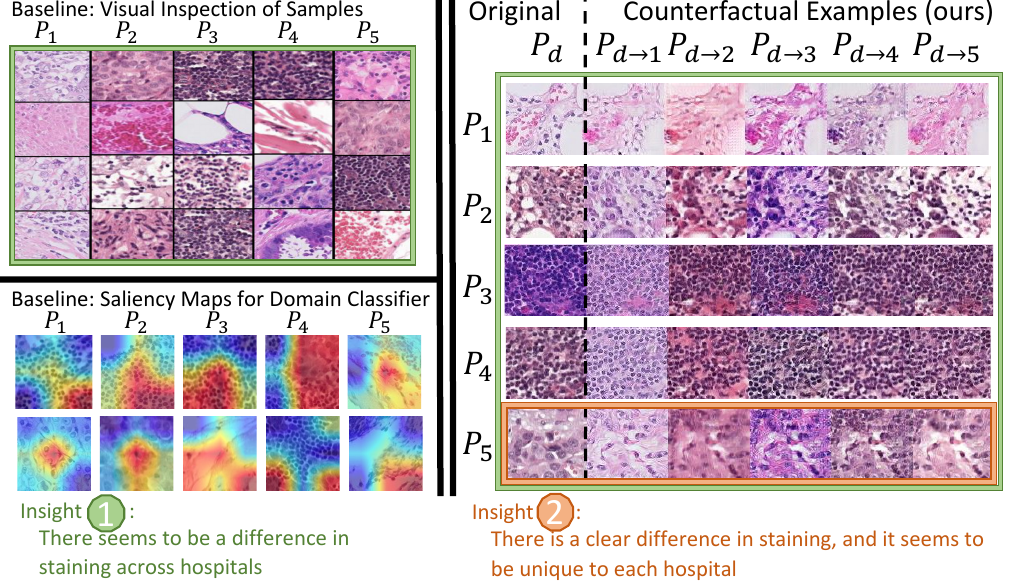}
    \vspace{-1.5 em}
    \caption{
    While the baseline method of unpaired samples (top-left) hints that there is a difference in staining (but is relatively unclear), our explanation approach (right) of showing paired counterfactual images translated between the hospital domains (represented as $P_1$, $P_2$, $\dots$) quickly leads to the stronger {\color{orange} \textbf{insight 2}} ---indeed the staining/coloring differs across the hospital domains \emph{and} the type of staining seems to be consistent and unique for each hospital.
    For the counterfactual examples, the $(i_{\text{row}},j_{\text{column}})$ pair represents the pushforward of a sample from domain $P_i$ onto the $P_j$ domain.
    Using Grad-CAM \cite{selvaraju2016grad} to explain a ResNet-50 \cite{he2016deep} domain classifier (bottom-left) does not lead to actionable insights.
    }   
    \label{fig:shift-counterfactuals}
    \vspace{-1.5 em}
\end{figure}

\paragraph{Explaining Shifts in H\&E Images Across Hospitals}
We apply this distribution counterfactual approach to the Camelyon17 dataset \cite{bandi2018detection} which is a real-world distribution shift dataset that consists of whole-slide histopathology images from five different hospitals.
We use the Stanford WILDS \cite{koh2021wilds} variant of the dataset which converts the whole-slide images into over 400 thousand patches.
Since each hospital has varying hematoxylin and eosin (H\&E) staining characteristics, this, among other batch effects, leads to heterogeneous image distributions across hospitals, as suggested by {\color{green} \textbf{insight 1}} in \autoref{fig:shift-counterfactuals}. 

To generate the counterfactual examples, we treat each hospital as a domain and train a StarGAN model \cite{choi2018stargan} to translate between each domain.
For training, we followed the original training approach seen in \cite{choi2018stargan}, with the exception that we perform no center cropping.
After training, we can generate distribution counterfactual examples by inputting a source image and the label of the target hospital domain to the model.
Counterfactual generation was done for all five hospitals and can be seen on the right-hand side of \autoref{fig:shift-counterfactuals}.
It can be seen that the distributional counterfactuals lead to {\color{orange} \textbf{insight 2}} ---that each hospital has a distinct level of staining that seems to be characteristic across samples from the same hospital.
For example, $P_1$ (hospital 1) consists of mostly light staining, and thus transporting to this domain usually involves lightening of the image.
Thus, if a practitioner is building a model which should generalize to slides from a new hospital within a network, {\color{orange} \textbf{insight 2}} confirms that controlling for different levels of staining (\eg color augmentations) is necessary in the training pipeline.
We further explore distinctly content-based counterfactuals (as opposed to style changes such as levels of staining) of an image using the CelebA dataset in \autoref{ssec:counterfactual-explanations-with-CelebA}.

\section{Discussion and Limitations}

\textbf{Choosing Between $k$-sparse vs. $k$-cluster Explanations}
We first highlight that distribution shift is a highly context-specific problem.
Thus, the explanation method will likely depend on the nature of the data (e.g., the data may contain natural subgroups or clusters).
If the sparsity or cluster structure is unknown, we suggest following the logical flow for method selection in \autoref{fig:main} to determine which shift explanation method to use for their specific context. 
In short, the $k$-sparse mappings are useful to allow an operator to see how specific \emph{features} changed from $\sourceD$ to $\targetD$, and the $k$-cluster mappings are useful to track how \emph{sub-groups of samples} changed under the distribution shift.

\textbf{Evaluating Shift Explanations} A primary challenge in developing distribution shift explanations is determining how to evaluate the efficacy of a given explanation in a given context.
Evaluating explanations is an active area of research \cite{robnik2018perturbation, molnar2020interpretable, doshi2017towards} with commonalities such as an explanation should be contrastive, succinct, should highlight abnormalities, and should have high fidelity \cite{molnar2020interpretable}.
For the case of distribution shift explanations, as this is a highly context-dependent problem (dependent on the data setting, task setting, and operator knowledge) and our approach is designed to tackle this problem in general, we do not have a general automated way of measuring whether a given explanation is indeed interpretable.
Instead, we provide a general contrastive method that supplies the PercentExplained (approximation of fidelity) and the adjustable $k$-level of sparse/cluster mappings (which trades off between succinctness and fidelity) but ultimately leaves the task of validating the explanation up to the operator.

\textbf{Potential Failure Cases} \hspace{0.25 em} 
Our explanations are meant as a diagnostic tool like most initial data analyses. 
While our explanations can provide actionable insights, they are one step in the diagnosis process and need to be investigated further before making a decision, especially in high-stakes scenarios. 
As with most explanation methods, the explanations may be incorrectly interpreted as being \emph{causal} while they likely only show correlations and many hidden confounders could exist.
While causal-oriented explanation methods are much more difficult to formulate and optimize, they could provide deeper insights than standard methods.
Because OT with squared Euclidean cost is the basis for several of our methods, this could cause a misleading explanation if the scaling of the dimensions does match the operator's expectations (more discussion below).
Another example failure case would be using $k$-cluster transport when no natural clusters exist. 
In practice, we noticed that StyleGAN can fail to recover large content-based shifts (as opposed to style-based shifts) such as the shift from “wearing hat” $\rightarrow$ “bald” in Celeb-A (as seen in \autoref{fig:celebA-shift-counterfactuals}).
This case is difficult to diagnose, but could be alleviated by using an Image-To-Image translation approach which does not have the style-based biases seen in methods like StyleGAN \cite{karras2019style}.

 \textbf{Motivation for using Optimal Transport and Euclidean cost} \hspace{0.25 em}
 Because there are many possible mappings between source and target distributions, any useful mapping will need to make some assumptions about the shift.
 We chose the OT assumption, which can be seen as a simplicity assumption (similar to Occam's Razor) because points are moved as little as possible.
 Additionally, the OT mapping is unique (even with just samples) and can be computed with fast known algorithms \citep{peyre2019Computational}.
 The squared Euclidean distance is a reasonable cost function when pointwise distances are semantically or causally meaningful (e.g., increasing the education of a person)--as is the case for many shift settings in this work.
However, squared Euclidean distance in the raw feature space might not be meaningful for some datasets such as raw pixel values of images.
For cases like these, a cost function other than the $\ell^2_2$ cost can be used, and we explore this in \autoref{ssec:interpretable-mappings-for-images} and in detail in \autoref{sec:appendix-explaining-shifts-in-images}. 
In those sections, we look at first applying a semantic-encoding function $g(\mathbf{x})$ which projects $\mathbf{x}$ to a semantically meaningful latent space and then calculate the transportation cost in this meaningful latent space.
In general, because OT algorithms can use any distance function, context-dependent cost functions could easily be used within our framework for improved interpretability.

\textbf{Future Directions For Shift Explanations} \hspace{0.25 em}
We believe developing new shift explanation maps and evaluation criteria for specific applications (e.g., explaining the results of biological experiments run with different initial conditions) is a rich area for future work.
Also, the PercentExplained metric does not provide information on specifically what is missing from the explanation, i.e., the missing information is a “known unknown”.
For image-based explanations, the explanation may fail to show certain domain changes that could mislead an operator, i.e., ``unknown unknowns''.
Methods to quantify and analyze these unknowns would improve the trustworthiness of shift explanations.
For further discussions of challenges with explaining distribution shifts (e.g., finding an interpretable latent space, approximations of Wasserstein distances in high dimensional regimes, etc.) we point the reader to \autoref{sec:limitations_and_challenges}.

\vspace{-0.5 em}
\section{Conclusion}
\vspace{-0.5 em}
In this paper, we introduced a framework for explaining distribution shifts using a transport map $T$ between a source and target distribution.
We constrained a relaxed form of optimal transport to theoretically define an intrinsically interpretable mapping $\TIT$ and introduced two interpretable transport methods: $k$-sparse and $k$-cluster transport.
We provided practical approaches to calculating a shift explanation, which allows us to use treat interpretability as a hyperparameter that can be adjusted based on a user's need and showed how our methods can help an operator investigate a distribution shift on real-world examples.
Both in \autoref{sec:experiments} and in \autoref{sec:simulated-and-simple-experiments}, we show the feasibility of our techniques on many different shift problems to both gain intuition for the different types of shift explanations and to show how our methods can help an operator investigate a distribution shift.
We hope our work suggests multiple natural extensions such as using trees as a feature-axis-aligned form of clustering or even other forms of interpretable sets.
Given our results and potential ways forward, we ultimately hope our framework lays the groundwork for providing more information to aid in investigations of distribution shift.

\textbf{Acknowledgements} This work was supported in part by ARL (W911NF-2020-221) and ONR (N00014-23-C-1016).

\bibliography{seank-ref}
\bibliographystyle{icml2023}

\newpage
\appendix
\onecolumn

\section{Proofs}
\label{sec:appendix-proofs}

\subsection{Proof that there are an infinite number of possible mappings between distributions}
\label{ssec:examples-of-infinite-mappings}
As stated in the introduction, given two distributions, there exist many possible mappings such that $\Tpush \sourceD = \targetD$ (it should be noted that here we are speaking of the general mapping problem, not the \emph{optimal} transport problem which can be shown  via Brenier's theorem \cite{peyre2019Computational} to have a unique matching for some cases).
For instance, given two isometric Gaussian distributions $\xvec \sim \mathcal{N}_1(\mathbf{\mu}_1, I)$, $\yvec \sim \mathcal{N}_2(\mathbf{\mu}_2, I)$, where $I$ is the Identity matrix, there exist an infinite number of $T$'s such that $T(\xvec) \sim \mathcal{N}_2$.
Specifically, any $T$ of the form: $T(\xvec) = \mathbf{\mu}_2 + R ( \xvec - \mathbf{\mu}_1 )$, where is $R$ is an arbitrary rotation matrix, will shift $\Tpush \mathcal{N}_1$ to have a mean of $\mathbf{\mu}_2$ and perfectly align the two distributions (since any rotation of an isometric Gaussian will still be an isometric Gaussian).

\subsection{Proof that practical interpretable transport objective is an upper bound of theoretic interpretable transport}
\label{ssec:emperical-IT-upper-bound}
First, recall our empirical approximation problem for finding $T$ where the second term is an approximation to the Wasserstein distance: 
\begin{equation}
    \label{eq:Emp-dist-from-OT-loss}
    \argmin_{T \in \Omega} \frac{1}{N} \sum_{i=1}^N c ( \xvec^{(i)}, T ( \xvec^{(i)} ) ) + 
    \lambda d ( T ( \xvec^{(i)} ), \TOT ( \xvec^{(i)} ) )
\end{equation}
where $\TOT$ is the optimal transport solution between our source and target domains with the given $c$ cost function.
The empirical average over samples can be viewed as an empirical expectation (which converges to the population expectation as $N$ approaches infinity).
Because the Wasserstein distance is well-defined for discrete distributions (like the empirical distribution) and continous distributions, we can simply prove that our approximation is an upper bound for any expectation (empirical or population-level) as follows:
\begin{align}
    W_2^2(P_{T(\xvec)}, P_Y) 
    &= \min_{T':T'_\sharp P_{T(\xvec)} = \targetD} \E_{\zvec \sim P_{T(\xvec)}} [d \left( \zvec, T'( \zvec) \right)] \label{eqn:by-definition} \\
    &= \min_{T':T'_\sharp P_{T(\xvec)} = \targetD} \E_{\xvec \sim \sourceD} [d \left( T(\xvec), T' \circ T(\xvec)  \right)] \label{eqn:change-of-expectation} \\
    &\leq \E_{\xvec \sim \sourceD} [d \left( T(\xvec), (T_{OT} \circ T^{-1}) \circ T(\xvec)  \right)] \label{eqn:one-speific-case}\\
    &= \E_{\xvec \sim \sourceD} [d \left( T(\xvec), T_{OT}(\xvec)  \right)] \,, \label{eqn:by-simplification}
\end{align}
where \autoref{eqn:by-definition} is by definition of Wasserstein distance, \autoref{eqn:change-of-expectation} is by a change of variables, \autoref{eqn:one-speific-case} is by taking $T' = T_{OT} \circ T^{-1}$ (which by construction satisfies the alignment constraint and which must be greater or equal to the minimum), and \autoref{eqn:by-simplification} is merely as simplification.
Note that if $T$ is the identity, then the inequality becomes an equality.
A similar proof could be used for any $W_p^p$ distance where $p \geq 1$.

\subsection{Proof that $k$-sparse truncated OT minimizes alignment upper bound}
In this section, we prove that the best possible alignment objective in terms of the upper bound on Wasserstein distance in \autoref{eq:emperical-interpretable-transport} is given by the truncated OT solution, i.e., the solution in the limit as $\lambda \to \infty$.
This solution is more akin to the constraint-based (i.e., non-Lagragian relaxation) of OT except replacing the alignment metric with the upper bound above.
While this enforces the best possible alignment and does not directly consider the transportation cost, the solution is relatively low cost because it based on the OT solution.
Additionally, it is the \emph{unique} solution to the problem as $\lambda \to \infty$.

We will now prove that it is the optimal and unique solution. First, let $\zvec = T(\xvec)$, $\zvec^{OT}= \TOT(\xvec)$, and  $\xvec \in \R^{N \times D}$.
If $d$ is the squared Euclidean distance and we restrict to mappings that only change dimensions in $\Aset$, then we can decompose the distance term as follows:
\begin{align}
    \sum_{i=1}^N d(\zvec_i, \zvec^{OT}_i) 
    = \sum_{i = 1}^N \sum_{j \in \Aset} \left( \zvec_{i,j} - \zvec^{OT}_{i,j} \right) ^2 + 
    \underbrace{\sum_{j \not\in \Aset} \left( \xvec_{i,j} - \zvec^{OT}_{i,j} \right) ^2}_{=\alpha_i \text{  ,  since constant w.r.t T}} 
    = \sum_{i = 1}^N \alpha_i + \sum_{j \in \Aset}  \left( \zvec_{i,j} - \zvec^{OT}_{i,j} \right) ^2  \,.
    \label{eqn:decompose-distance}
\end{align}
Notice that the sum of squares corresponding to $\Aset$ is dependent on the mapping while the others are a constant w.r.t. $T$ because $T$ cannot modify any dimensions $j \not\in \Aset$.
Given this, we now choose our solution to $k$-sparse optimal transport as given in the paper: 
\begin{align}
    \forall j, [T(\xvec)]_j = \left \{ 
    \begin{array}{ll}
    [\TOT(\xvec)]_j, & \text{if} \,\, j \in \Aset \\
    x_j, & \text{if} \,\, j \not\in \Aset \\
    \end{array}
    \right.
    \label{eqn:optimal-T-solution}
\end{align}
where $\Aset$ is the active set of $k$ dimensions which our $k$-sparse map $T$ can move points.
With this solution, we arrive at the following:
\begin{align*}
\sum_{i=1}^N d(\zvec_i, \zvec^{OT}_i)
= \sum_{i = 1}^N \alpha_i + \sum_{j \in \Aset} \left( \zvec_{i,j} - \zvec^{OT}_{i,j} \right) ^2 
=\sum_{i = 1}^N \alpha_i + \sum_{j \in \Aset} \sum_{i = 1}^N \left( \zvec^{OT}_{i,j} - \zvec^{OT}_{i,j} \right) ^2 
= \sum_{i=1}^N \alpha_i \,,
\end{align*}
where the $\alpha_i$ are positive constants that cannot be reduced by $T$.
Therefore, this is indeed the optimal solution to our empirical interpretable transport problem with the alignment approximation as in \autoref{eq:emperical-interpretable-transport}.
This can easily be extended to show that the optimal active set for this case is the one that minimizes $\sum_{i=1}^N \alpha_i$, thus the active set should be the $k$ dimensions which have the largest squared difference between $\xvec$ and $\zvec^{OT}$.

To prove uniqueness, we use proof by contradiction. Suppose there exists another optimal solution $T'$ that is distinct from the optimal $T$ given in \autoref{eqn:optimal-T-solution}.
This would mean that there exists a pair $(\xvec, j)$ such that $[T'(\xvec)]_j \neq [T(\xvec)]_j = [\TOT(\xvec)]_j = z_j^{OT}$.
However, this would mean that the corresponding term in the summation would be non-zero, i.e., $(z_{j} - z_j^{OT})^2 > 0$.
But this would mean that the overall distance function is greater than the sum of the constants yielding a contradiction to the hypothesis that there could exist another solution.
Therefore, the $k$-sparse solution that optimizes alignment is unique.

\subsection{Proof that $k$-mean shift is the $k$-vector shift that gives the best alignment}

Similar to the previous proof, we consider the solution to \autoref{eq:emperical-interpretable-transport} when $\lambda \to \infty$, i.e., the optimal alignment solution, but restrict ourselves to the space of vector maps $\Omega_{vector}$.
First, we recall the definition of $\Omega_{vector}$:
\begin{align}
    \kIntSet_{vector} = \{ T : T(\xvec) = \xvec + \tilde{\delta} \},  
    \quad \text{where} \quad
    \tilde{\delta}_j = \left \{ 
    \begin{array}{ll}
       \delta_j,  & \text{if}\,\, j \in \Aset \\
       0 & \text{if}\,\, j \not\in \Aset 
    \end{array}
    \right. \,,
\end{align}
where $\delta_j$ for $j\in\Aset$ are the only learnable parameters.
Given this, we decompose the sum of distances similar to the previous proof:
\begin{align}
    \sum_{i=1}^N d(\zvec_i, \zvec^{OT}_i) 
    &= \sum_{i = 1}^N \alpha_i + \sum_{j \in \Aset}  \left( \zvec_{i,j} - \zvec^{OT}_{i,j} \right) ^2  
    \label{eqn:prior-decompose} \\
    &= \sum_{i = 1}^N \alpha_i + \sum_{j \in \Aset}  \left( (\xvec_{i,j} + \delta_j) - \zvec^{OT}_{i,j} \right) ^2  
    \label{eqn:by-def-of-T}
\end{align}
where \autoref{eqn:prior-decompose} is from \autoref{eqn:decompose-distance} and \autoref{eqn:by-def-of-T} is by the definition of $\kIntSet$.
Because this is a convex function that decomposes over each coordinate, we can take the derivative and set to zero to solve:
\begin{align}
    \frac{d}{d\delta_j} \left(\sum_{i=1}^N d(\zvec_i, \zvec^{OT}_i)  \right) 
    = \frac{d}{d\delta_j} \sum_{i=1}^N \left( (\xvec_{i,j} + \delta_j) - \zvec^{OT}_{i,j} \right) ^2
    = 2\sum_{i=1}^N \left( (\xvec_{i,j} + \delta_j) - \zvec^{OT}_{i,j} \right)
    = 2 \left( N\delta_j + \sum_{i=1}^N \xvec_{i,j}  - \zvec^{OT}_{i,j} \right) \,,
\end{align}
where the first equals is by \autoref{eqn:by-def-of-T} and noticing that other terms do constants w.r.t. $\delta_j$ and the rest is simple calculus.
Solving this for $\delta_j$ yields the following simple solution:
\begin{align}
    \delta_j 
    = \frac{1}{N}\sum_{i=1}^N  \zvec^{OT}_{i,j} - \xvec_{i,j}
    = \mu_j^{\zvec^{OT}} - \mu_j^{\text{src}}
    = \mu_j^{\text{tgt}} - \mu_j^{\text{src}}
\end{align}
where the second is just by definition of the mean, and the last is by noticing that the mean of the projected OT samples is equal to the man of the target samples since the projected samples will match the target dataset by construction of the OT solution.
This solution matches the one in the main paper which we recall here:
\begin{align}
    \forall j, [T(\xvec)]_j = \left \{ 
    \begin{array}{ll}
    x_j + (\mu_{j}^{\text{tgt}} - \mu_{j}^{\text{src}}), & \text{if} \,\, j \in \Aset \\
    x_j, & \text{if} \,\, j \not\in \Aset \\
    \end{array}
    \right. \,,
\end{align}
Thus showing the optimal delta vector to minimize $k$-vector transport is exactly the $k$-sparse mean shift solution.

\section{Challenges of Explaining Distribution Shifts and Limitations of Our Method}
\label{sec:limitations_and_challenges}
Distribution shift is a ubiquitous and quite challenging problem.
Thus, we believe discussing the challenges of this problem and the limitations of our solution should aid in advancements in this area of explaining distribution shifts.

As mentioned in the main body, as distribution shifts can take many forms, trying to explain a distribution shift is a highly context-dependent problem (i.e., dependent on the data setting, task setting, and operator knowledge).
Thus, a primary challenge in developing distribution shift explanations is determining how to evaluate whether a given explanation is valid for a given context.
In this work, we hope to introduce the problem of explaining distribution shifts \emph{in general} (i.e. not with a specific task nor setting in mind), therefore we do not have an automated way of measuring whether a given explanation is indeed interpretable.
Evaluating explanations is an active area of research \cite{robnik2018perturbation, molnar2020interpretable, doshi2017towards} with commonalities such as an explanation should be contrastive, succinct, should highlight abnormalities, and should have high fidelity.
Instead, we introduce a proxy method that supplies the operator with the PercentExplained and the adjustable $k$-level of sparse/cluster mappings but leaves the task of validating the explanation up to the operator.
We believe developing new shift explanation maps and criteria for specific applications (e.g., explaining the results of experiments run with different initial conditions) is a rich area for future work.

Explaining distribution shifts becomes more difficult when the original data is not interpretable. 
This typically can take two forms: 
1) the raw data \emph{features} are uninterpretable but the samples are interpretable (e.g., a sample from the CelebA dataset \cite{liu2015faceattributes} is interpretable but the pixel-level features are not)
or 2) when both the raw data features and samples are uninterpretable (e.g., raw experimental outputs from material science simulations).
In the first case, one can use the set of counterfactual pairs method outlined in \autoref{ssec:counterfactual-examples}  (see \autoref{fig:celebA-shift-counterfactuals} for examples with CelebA), however, as mentioned in the main paper, this is less sample efficient than an interpretable transport map.
For the second case, if the original features are not interpretable, one must first find an interpretable latent feature space -- which is a challenging problem by itself.
As seen in \autoref{fig:high-dim-cluster-explanations}, it is possible to solve for a semantic latent space and solve interpretable transport maps within the latent space, in this case, the latent space of a VAE model. 
However, if the meaningful latent features are not extracted, then any transport map within this latent space will be meaningless.
In the case of \autoref{fig:high-dim-cluster-explanations}, the 3-cluster explanation is likely only interpretable because we know the ground truth and thus know what to look for. 
As such, this is still an open problem and one we hope future work can improve on.

Additionally, while the PercentExplained metric shows the fidelity of an explanation (i.e. how aligned $T_\sharp (P_{src})$ and $P_{tgt}$ are), we do not have a method of knowing specifically \emph{what} is missing from the explanation. 
This missing part of the explanation can be considered a “known unknown”.
For example, if a given $T$ has a PercentExplained of 85\%, we know how much is missing, but we do not know what information is contained in the missing 15\%.
Similarly, when trying to explain an image-based distribution shift with large differences in content (e.g., a dataset with blonde humans and a dataset with bald humans), leveraging existing style transfer architectures (where one wishes to only change the style of an image while retaining as much of the original content as possible) to generate distributional counterfactuals can lead to misleading explanations.
This is because explaining image-based distribution shifts might require large changes in content (such as removing head hair from an image), which most style-transfer models are biased against doing.
As an example, \autoref{fig:celebA-shift-counterfactuals} shows an experiment that translates between five CelebA domains (blond hair, brown hair, wearing hat, bangs, bald).
It can be seen that the StarGAN model can successfully translate between stylistic differences such as “blond hair” $\rightarrow$ “brown hair” but is unable to make significant content changes such as “bangs” $\rightarrow$ “bald”.

The above issues are mainly problems that affect distribution shift explanations in general, but below are issues specific to our shift explanation method (or any method which similarly uses empirical OT).
Since we rely on the empirical OT solution for the sparse and cluster transport (and the percent explained metric), the weaknesses of empirical OT are also inherited.
For example, empirical OT, even using the Sinkhorn algorithm with entropic regularization, scales at least quadratically in the number of samples \cite{cuturi2013sinkhorn}.
Thus, this is only practical for thousands of samples.
Furthermore, empirical OT is known to poorly approximate the true population-level OT in high dimensions although entropic regularization can reduce this problem \cite{genevay2019sample}.
Finally, empirical OT does not provide maps for new test points.
Some of these problems could be alleviated by using recent Wasserstein-2 approximations to optimal maps via gradients of input-convex neural networks based on the dual form of Wasserstein-2 distance \cite{korotin2019wasserstein, makkuva2020optimal}.
Additionally, when using $k$-cluster maps, the clusters are not guaranteed to be significant (i.e. it might be indiscernible what makes this cluster different than another cluster), and thus if using $k$-cluster maps on datasets that do not have natural significant clusters (e.g., $\sourceD \sim $Uniform$(0, 1)$,  $\targetD \sim $Uniform$(1, 2)$) an operator might waste time looking for significance where there is none. 
While this cannot be avoided in general, using a clustering method that is either specifically designed for finding interpretable clusters \cite{fraiman2013interpretable, bertsimas2021interpretable} or one which directly optimizes the objective in interpretable transport equation \autoref{eq:Interpretable-Transport} might lead to easier to explanations which are easier to interpret or validate.

\section{Experiments on Known Shifts}
\label{sec:simulated-and-simple-experiments}
Here we present additional results on simulated experiments as well as an experiment on UCI ``Breast Cancer Wisconsin (Original)'' dataset \cite{mangasarian1990cancer}.
Our goal is to illuminate when to use the different sets of interpretable transport, and how the explanations can be interpreted, where in this case, a ground truth explanation is known.
\footnote{Code to recreate all experiments can be found at \href{https://github.com/inouye-lab/explaining-distribution-shifts}{https://github.com/inouye-lab/explaining-distribution-shifts}.}

\begin{figure}[!ht]
    \centering
    \includegraphics[width=0.7\textwidth]{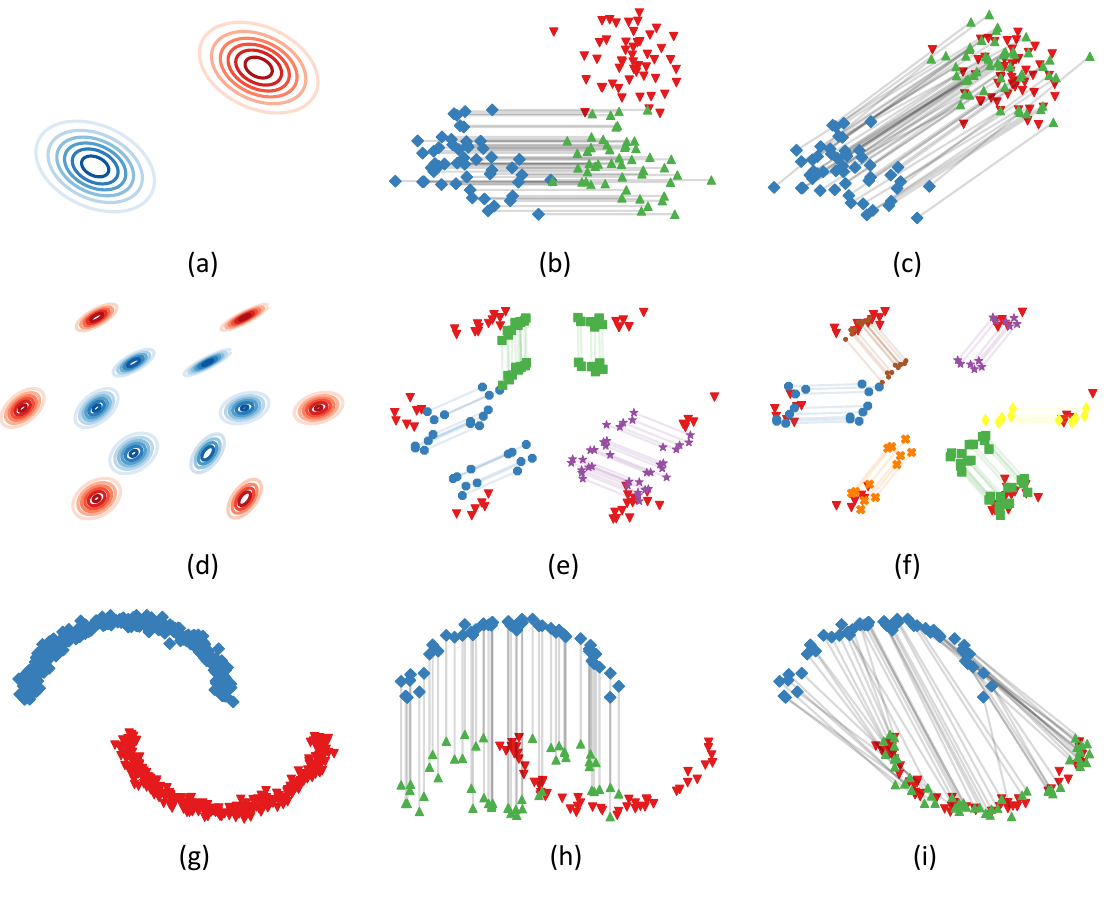}
    \caption{Three toy dataset shift examples showing the advantages of the different shift explanation methods, where a mean shift between Gaussians (top row) can be easily explained using $k$-sparse vector shifts, a varying mean shift across mixture components of a Gaussian mixture model (middle row) is best explained using $k$-sparse transport maps, while a complex shift (bottom row) requires a complex feature-wise mapping, such as $k$-sparse optimal transport, which maximally aligns the distributions as it can perform conditional transport mappings for each sample (as seen by the differing vertical shifts in (h) depending on where the blue sample lies on the horizontal axis), at the expense of interpretability.
    Each example shows three levels of decreasing interpretability, where the leftmost column shows the original shift (which has maximal interpretability since $k=0$) from source (blue diamonds) to target (red down arrows), and the rightmost column shows a shift with near-perfect fidelity.}
    \label{fig:toy-problems-full}
\end{figure}

\subsection{Simulated Experiments}
In this section we study three toy shift problems: a mean shift between two, otherwise identical, Gaussian distributions, a Gaussian mixture model where each mixture component has a different mean shift, and a flipped and shifted half-moon dataset, as seen in figures (a), (d), and (g) in \autoref{fig:toy-problems-full}.

The first case is a mean shift between two, otherwise identical, Gaussian distributions can be easily explained using $k$-sparse mean shift (as well as vanilla mean shift).
We first calculate the OT mapping $\TOT$ between the two Gaussian distributions, which has a closed form solution of $\TOT(\xvec) = \mu_{tgt} + A ( \xvec - \mu_{src} )$, where $A$ is a matrix that can be seen as a conversion between the source and target covariance matrices, and because the covariance matrices are identical, A is the identity.

The second toy example of distribution shift is a shifted Gaussian mixture model which represents a case where groups within a distribution shift in different ways.
An example of this type of shift could be explaining the change in immune response data across patients given different forms of treatment for a disease.
Looking at (d) in \autoref{fig:toy-problems-full}, it is clear that sparse feature transport will not easily explain this shift.
Instead, we turn to cluster-based explanations, where we first find $k$ paired clusters and attempt to show how these shift from $\sourceD$ to $\targetD$.
Following the mean-shift transport of paired clusters approach outlined in \autoref{ssect:cluster-based-shift}, the $k=3$ case as seen in the Appendix shows that three clusters can sufficiently approximate the shift by averaging the shift between similar groups.
If a more faithful explanation is required, (f) of \autoref{fig:toy-problems-full} shows that increasing $k$ to 6 clusters can recover the full shift, \ie PercentExplained=100, at the expense of being less interpretable (which is especially true in a real-world case where the number of dimensions might be large).

The half-moon example, figure (g) in \autoref{fig:toy-problems-full}, shows a case where a complex feature-wise dependency change has occurred.
This example is likely best explained via feature-wise movement, so will use $k$-sparse transport.
If we follow the approach in \autoref{ssec:sparse-explanation-maps} with our interpretable set as the $\Omega^{(k)}$ and let $k=1$, we get a mapping that is interpretable, but has poor alignment (see Figure (h) in \autoref{fig:toy-problems-full}).
For this example, we can possibly reject this explanation due to a poor PercentExplained.
With the understanding that this shift is not explainable via just one feature, we can instead use a $k=2$-sparse OT solution.
The $k=2$ case can be seen in (i) of \autoref{fig:toy-problems-full} which shows that this approach aligns the distributions perfectly, at the expense of interpretability.

\subsection{Explaining Shift in Wisconsin Breast Cancer Dataset} 

\begin{figure}[!h]
    \centering
    \includegraphics[width=\textwidth]{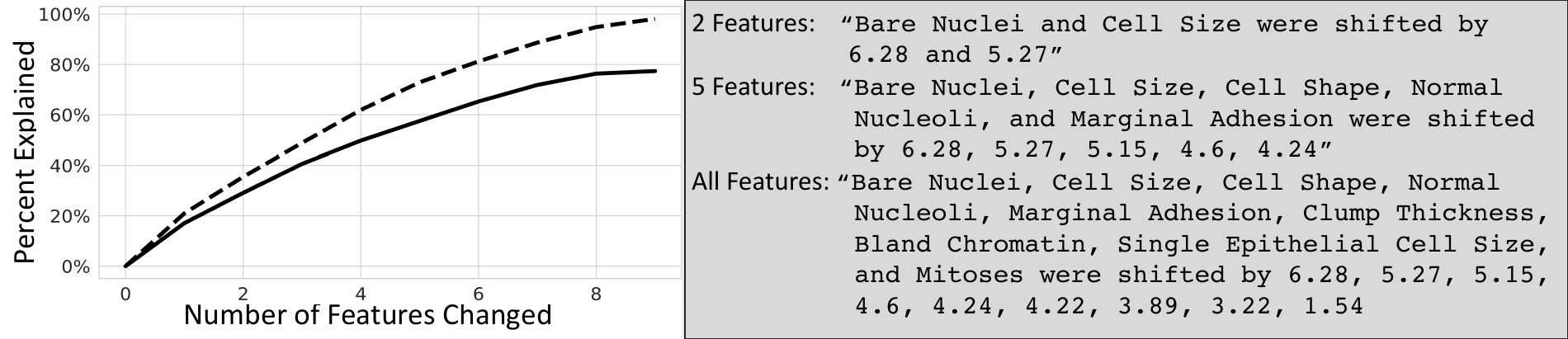}
    \caption{
    A comparison of the performance of $k$-sparse mean shift explanations (solid line) and $k$-sparse optimal transport explanations (dashed line) when explaining the shift from the benign tumor samples to malignant tumor samples for the UCI Wisconsin Breast Cancer dataset.
    On the right are example explanations a human operator would see as they change the level of interpretability during $k$-sparse mean shift explanations (where ``All Features'' is the baseline full mean shift explanation).}
    \label{fig:wisconsin-results}
\end{figure}

This tabular dataset consists of tumor samples collected by \citet{mangasarian1990cancer} where each sample is described using nine features which are normalized to integers from $[0, 10]$.
We split the dataset along the class dimension and set $\sourceD$ to be the 443 benign tumors and $\targetD$ as the 239 malignant samples.
To explain the shift, we used two forms of $k$-sparse transport, the first being $k$-sparse mean transport and the second being $k$-sparse optimal transport.
The left of \autoref{fig:wisconsin-results} shows that the $k$-sparse mean shift explanation is sufficient for capturing the 50\% of the shift between $\sourceD$ and $\targetD$ using only four features, and nearly 80\% of the shift with all 9 features.
However, if an analyst requires a more faithful mapping, they can use the $k$-sparse OT explanation which can recover the full shift, at the expense of the interpretability.
The right of \autoref{fig:wisconsin-results} shows example explanations that an analyst can use along with their context-specific expertise for determining the main differences between the different tumors they are studying.

\subsection{Counterfactual Example Experiment to Explain a Multi-MNIST shift } \label{ssec:toy-counterfactual-examples}

\begin{figure}[!h]
    \centering
    \includegraphics[width=0.8\textwidth]{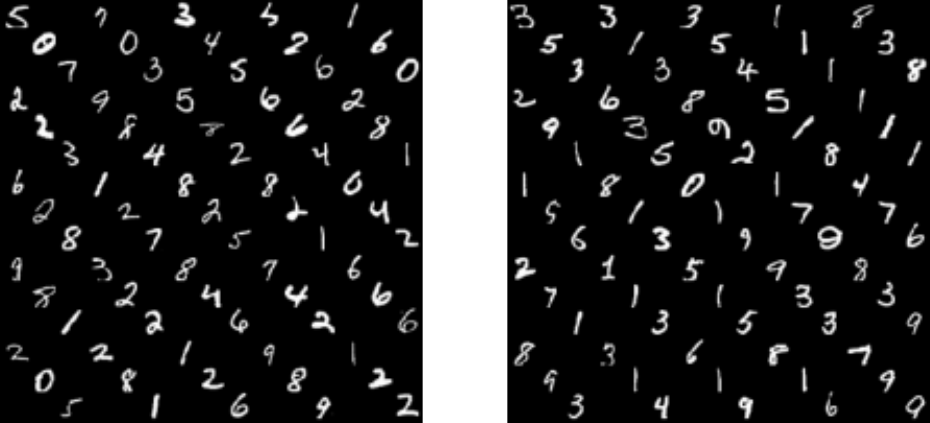}
    \caption{A grid of 25 raw samples from each domain (left is $\sourceD$ and right is $\targetD$). 
    Even for the relatively simple shift seen in the Shifted Multi-MNIST dataset, it may be hard to tell what is different between the two distributions by just looking at samples (without knowing the oracle shift).
    Each sample in this dataset contains three MNIST digits along a diagonal and the domain label corresponds to the evenness of the middle MNIST digit (where $\sourceD$ contains even middle digits and $\targetD$ contains odd middle digits).
    }
    \label{fig:multi-mnist-raw-dataset-examples}
\end{figure}

As mentioned in \autoref{ssec:counterfactual-examples}, image-based shifts can be explained by supplying an operator with a set of distributional counterfactual images with the notion that the operator would resolve which semantic features are distribution-specific.
Here we provide a toy experiment (as opposed to the real-world experiment seen in \autoref{ssec:counterfactual-examples}) to illustrate the power of distributional counterfactual examples.
To do this, we apply the distributional counterfactual example approach to a Multi-MNIST dataset where each sample consists of a row of three randomly selected MNIST digits \cite{deng2012mnist} and is split such that $\sourceD$ consists of all samples where the middle digit is even and zero and $\targetD$  is all samples where the middle digit is odd, as seen in \autoref{fig:multi-mnist-raw-dataset-examples}.

\begin{algorithm}
\caption{Generating distributional counterfactuals using DIVA}
\label{alg:counter-factual-generation}
\begin{algorithmic}
\STATE \textbf{Input:} $\xvec_1 \sim D_1$, $\xvec_2 \sim D_2$, model
\STATE $z_{y_1}, z_{d_1}, z_{{residual}_1} \gets $ model.encode$(\xvec_1)$
\STATE $z_{y_2}, z_{d_2}, z_{{residual}_2} \gets $ model.encode$(\xvec_2)$
\STATE $\hat{\xvec}_{1\rightarrow2} \gets $ model.decode($z_{y_1}, z_{d_2}, z_{{residual}_1}$)
\STATE $\hat{\xvec}_{2\rightarrow1} \gets $ model.decode($z_{y_2}, z_{d_1}, z_{{residual}_2}$)
\STATE \textbf{Output:} $\hat{\xvec}_{1\rightarrow2}$, $\hat{\xvec}_{2\rightarrow1}$
\end{algorithmic}
\end{algorithm}

\begin{figure}[H]
    \centering    
    \includegraphics[width=\linewidth]{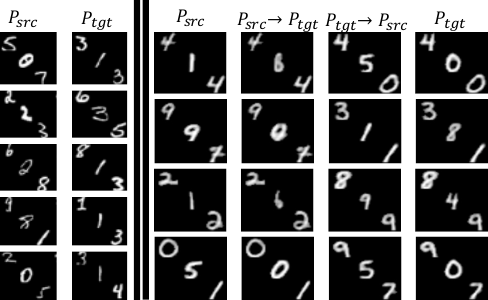}
    \caption{
    A comparison of the baseline grid of unpaired source and target samples (left) and counterfactual pairs (right) which show how counterfactual examples can highlight the difference between the two distributions.
    For each image, the top left digit represents the class label, the middle digit represents the distribution label (where $\sourceD$ only contains even digits and zero and $\targetD$ has odd digits), and the bottom right digit is noise information and is randomly chosen.
   The second, third columns show the counterfactuals from $\sourceD \rightarrow \targetD$ and $\targetD \rightarrow \sourceD$, respectively.
   Hence we can see under the push forward of each image the ``evenness'' of the domain digit changes while the class and noise digits remain unchanged.
   }
    \label{fig:toy-shift-counterfactuals}
\end{figure}

To generate the counterfactual examples, we use a Domain Invariant Variational Autoencoder (DIVA) \cite{ilse2020diva}, which is designed to have three independent latent spaces: one for class information, one for domain-specific information (or in this case, distribution-specific information), and one for any residual information.
We trained DIVA on the Shifted Multi-MNIST dataset for 600 epochs with a KL-$\beta$ value of 10 and latent dimension of 64 for each of the three sub-spaces.
Then, for each image counterfactual, we sampled one image from the source and one image from the target and encoded each image into three latent vectors: $z_{y}$, $z_{d}$, and $z_{residual}$.
The latent encoding $z_d$ was then ``swapped'' between the two encoded images, and the resulting latent vector set was decoded to produce the counterfactual for each image.
This process is detailed in Algorithm 2 below.
The resulting counterfactuals can be seen in \autoref{fig:toy-shift-counterfactuals} where the middle digit maps from the source (i.e., odd digits) to the target (i.e., even digits) and vice versa while keeping the other content unchanged (i.e., the top and bottom digits).

\subsection{Using StarGAN to Explain Distribution Shifts in CelebA}
\label{ssec:counterfactual-explanations-with-CelebA}

Here we apply the distributional counterfactual approach seen in \autoref{ssec:counterfactual-examples} to the CelebA dataset \cite{liu2015faceattributes}, which contains over 200K images of celebrities, each with 40 attribute annotations.
We split the original dataset into 5 related sets, $P_1$=``blonde hair'', $P_2$=``brunette hair'', $P_3$=``wearing hat'', $P_4$=``bangs'', $P_5$=``bald''.
These five sets were chosen as they are related concepts (all related to hair) yet mostly visually distinct.
Although there are images with overlapping attributes, such as a blonde/brunette person with bangs, these are rare and naturally occurring, thus they were not excluded.

\begin{figure}[H]
    \centering    
    \includegraphics[width=\linewidth]{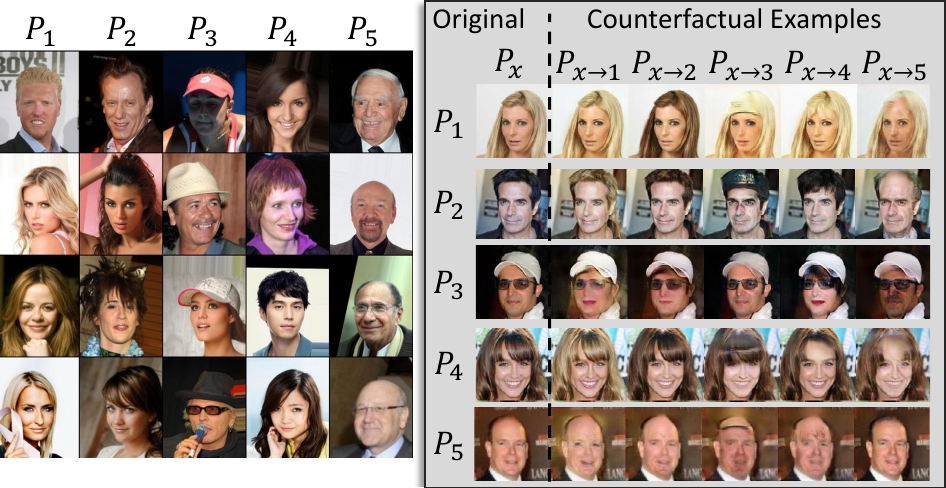}
    \caption{
    StarGAN is able to adequately translate between distributions with similar content but different style (\eg $P_1 \rightarrow P_2$), however, when transporting between distributions with different content (\eg "no hat" $\rightarrow P_3$) the I2I model is unable to properly capture the shift.
    This is likely due to the model being biased to only change the \emph{style} of the image, while maintaining as much \emph{content} as possible.
    The figure breakdown is similar to \autoref{fig:shift-counterfactuals} with the baseline method of unpaired samples on the left and paired counterfactual images on the right, where here $P_1$=``blonde hair'', $P_2$=``brunette hair'', $P_3$=``wearing hat'', $P_4$=``bangs'', $P_5$=``bald''.
    }
    \label{fig:celebA-shift-counterfactuals}
\end{figure}

We trained a StarGAN model \cite{choi2018stargan} to generate distributional counterfactuals following the same approach seen in \autoref{ssec:counterfactual-examples}.
The result of this process can be seen in \autoref{fig:celebA-shift-counterfactuals}, where we can see the model successfully translating ``stylistic'' parts of the image such as hair color.
However, the model is unable to translate between distributions with larger differences in ``content'' such as removing hair when translating to ``bald''.
This highlights a difference between I2I tasks such as style transfer (where one wishes to bias a model to only change the style of an image while retaining as much of the original content as possible) the mappings required for explaining image-based distribution shifts, which might require large changes in content (such as adding a hat to an image). 

\section{Explaining Shifts in Images via High-Dimensional Interpretable Transportation Maps}
\label{sec:appendix-explaining-shifts-in-images}
If $\xvec$ is an image with domain $\R^{D>>1}$, then any non-trivial transportation map in this space is likely to be hard to optimize for as well as uninterpretable. 
However, if $\sourceD, \targetD$ can be expressed on some \emph{interpretable} lower dimensional manifold which is learned by some manifold-invertible function $g: \R^D \rightarrow \R^{D'}$ where $D < D'$, we can project $\sourceD, \targetD$ onto this latent space and solve for an interpretable mapping such that it aligns the distributions in the latent space, $P_{T \left( g(\xvec) \right)} \approx P_{g(\yvec)}$.
Note, in practice, an encoder-decoder with an interpretable latent space can be used for $g$, however, requiring $g$ to be exactly invertible allows for mathematical simplifications, which we will see later.
For explainability purposes, we can use $g^{-1}$ to re-project $T \left( g(\xvec) \right)$ back to $\R^D$ in order to display the transported image to an operator.
With this, we can define our set of high dimensional interpretable transport maps: $\HighDIntSet \coloneq \left\{ T : T= g^{-1}\left( \tilde{T}\left( g(\xvec) \right) \right), \tilde{T} \in \kIntSet, g \in \mathcal{I} \right\}$ where $\kIntSet$ is the set of $k$-interpretable mappings (e.g., $k$-sparse or $k$-cluster maps) and $\mathcal{I}$ is the set of invertible functions with an interpretable (\ie semantically meaningful) latent space. 

Looking at our interpretable transport problem:
\begin{equation}
    \label{eq:appendix-interpretable-transport}
    \argmin_{T \in \HighDIntSet} 
    \E_{\sourceD} \left[ c(\xvec, T(\xvec) ) \right]
    + \lambda \phi (P_{T(\xvec))}, P_{\yvec})
\end{equation}

Although our transport is now happening in a semantically meaningful space, our transportation cost is still happening in the original raw pixel space.
This is undesirable since we want a transport cost that penalizes large semantic movements, even if the true change in the pixel space is small (\eg a change from ``dachshund'' to ``hot dog'' would be a large semantic movement).
We can take a similar approach as before and instead calculate our transportation cost in the $g$ space.
This logic can similarly be applied to our divergence function $\phi$ (especially if $\phi$ is the Wasserstein distance, since this term can be seen as the residual shift not explained by $T$).
Thus, calculating our cost and alignment functions within the latent space gives us:
\begin{equation}
    \argmin_{g \in \mathcal{I}, \tilde{T} \in \kIntSet} 
    \E_{\sourceD} \left[ c\left( g(\xvec), \tilde{T}\left( g(\xvec) \right) \right) \right]
    + \lambda \phi (P_{\tilde{T}(g(\xvec))}, P_{g(\yvec)})
\end{equation}

This formulation has a critical problem however.
Since we are jointly learning our representation $g$ and our transport map $T$, a trivial solution for the above minimization is for $g$ to map each point to an arbitrarily small space such that the distance between any two points $c(g(\xvec), g(\yvec))\approx 0$, thus giving us a near zero cost regardless of how ``far'' we move points.
To avoid this, we can use pre-defined image representation function $h$, e.g., the latter layers in Inception V3, and calculate pseudo-distances between transported images in this space.
Because $h$ expects an image as an input, we can utilize the invertibility of $g$ and perform our cost calculation as: $c\left( h(\xvec), h \left( g^{-1} \left( \tilde{T}\left( g(\xvec) \right) \right) \right) \right)$, or more simply, $c_h \left( \xvec,  T(\xvec) \right)$, where $T=g^{-1} \left( \tilde{T}\left( g(\xvec) \right) \right)$.
Similar to the previous equation, we also apply this $h$ pseudo-distance to our divergence function to get $\phi_h$.
With this approach, we can still use $g$ to jointly learn a semantic representation which is specific to our source and target domains (unlike $h$ which is trained on images in general, \eg ImageNet) and an interpretable transport map $\tilde{T}$ within $g$'s latent space.
This gives us:
\begin{equation}
    \label{eq:appendix-HIT}
    \argmin_{g \in \mathcal{I}, T \in \IntSet}
    \E_{\sourceD} \left[ c_h \left( \xvec,  T(\xvec) \right) \right]
    + \lambda \phi_h (P_{T(\xvec)}, P_{\yvec})
\end{equation}

While the above equation is an ideal approach, it can be difficult to use standard gradient approaches to optimize over in practice due it being a joint optimization problem and any gradient information having to first pass through $h$ which could be a large neural network.
To simplify this, we can optimize $\tilde{T}$ and $g$ separately.
With this, we can first find a $g$ which properly encodes our source and target distributions into a semantically meaningful latent space, and then find the best interpretable transport to align the distributions in the fixed latent space.
The problem can be even further simplified by setting the pre-trained image representation function $h$ to be equal to the pretrained $g$, since the disjoint learning of $g$ and $\tilde{T}$ removes the shrinking cost problem.
By setting $h \coloneqq g$, we can see that $c \left ( h(\xvec), h \circ g^{-1} \circ \tilde{T} \circ g (\xvec)  \right ) = c \left( g(\xvec), \tilde{T} \circ g(\xvec) \right) = c_g (\xvec, \tilde{T}(\xvec))$, which simplifies \autoref{eq:appendix-HIT} back to our interpretable transport problem, \autoref{eq:appendix-interpretable-transport}, where $g$ is treated as a pre-processing step on the input images:

\begin{equation}
    \label{eq:simplified-THIT}
    \argmin_{T \in \IntSet} 
    \E_{\sourceD} \left[ c( g(\xvec), g \left( T(\xvec) ) \right) \right]
    + \lambda \phi_g (P_{T(\xvec))}, P_{\yvec})
\end{equation}

Another way to simplify \autoref{eq:appendix-HIT} is to relax the constraint that $g$ is manifold-invertible and instead use a pseudo-invertible function such as an encoder $g$ and decoder $g^+$ structure  where $g^+$ is a pseudo-inverse to $g$ such that $g^+( g(\xvec) ) \approx \xvec$. This gives us:

\begin{equation}
\begin{aligned}
    \argmin_{\tilde{T} \in \tildeIntSet, g, g^+}
    \E_{\sourceD} \left[ c_h \left( \xvec,  g^+( \tilde{T}( g(\xvec) ) ) \right) \right]
    &+ \lambda_{Fid} ~ \phi_h (P_{g^+( \tilde{T}( g(\xvec) ) )}, P_{\yvec}) \\
    &+ \lambda_{Recon} ~ \E_{\frac{1}{2} \sourceD + \frac{1}{2} \targetD} \left[  L \left( \xvec, g^+( \tilde{T}( g(\xvec) ) ) \right) \right]
\end{aligned}
\end{equation}

where $L(\xvec, \cdot)$ is some reconstructive-loss function.

\subsection{Explaining a Colorized-MNIST shift via High-dimensional Interpretable Transport}
In this section we present a preliminary experiment showing the validity of our framework for explaining high-dimensional shifts.
The experiment consists of using $k$-cluster maps to explain a shift in a colorized-version of MNIST, where the source environment is yellow/light red digits with a light grayscale background color (\ie light gray) and the target environment consists of darker red digits and/or a darker grayscale background colors.
Like the lower dimensional experiments before, our goal is to test our method on a shift where the ground truth is known and thus the explanation can validated against.
We follow the framework presented in \autoref{eq:simplified-THIT},  where the fixed $g$ is a semi-supervised VAE \cite{siddharth2017learning} which is trained on a concatenation of $\sourceD$ and $\targetD$.
Our results show that $k$-cluster transport can capture the shift and explain the shift, however, we suspect the given explanation is interpretable because the ground truth is already known.
Our hope is that future work will improve upon this framework by better finding a latent space which is interpretable and disentangled, leading to better latent mappings, and thus better high-dimensional shift explanations.

\begin{figure}[!ht]
    \centering
    \includegraphics[width=\textwidth]{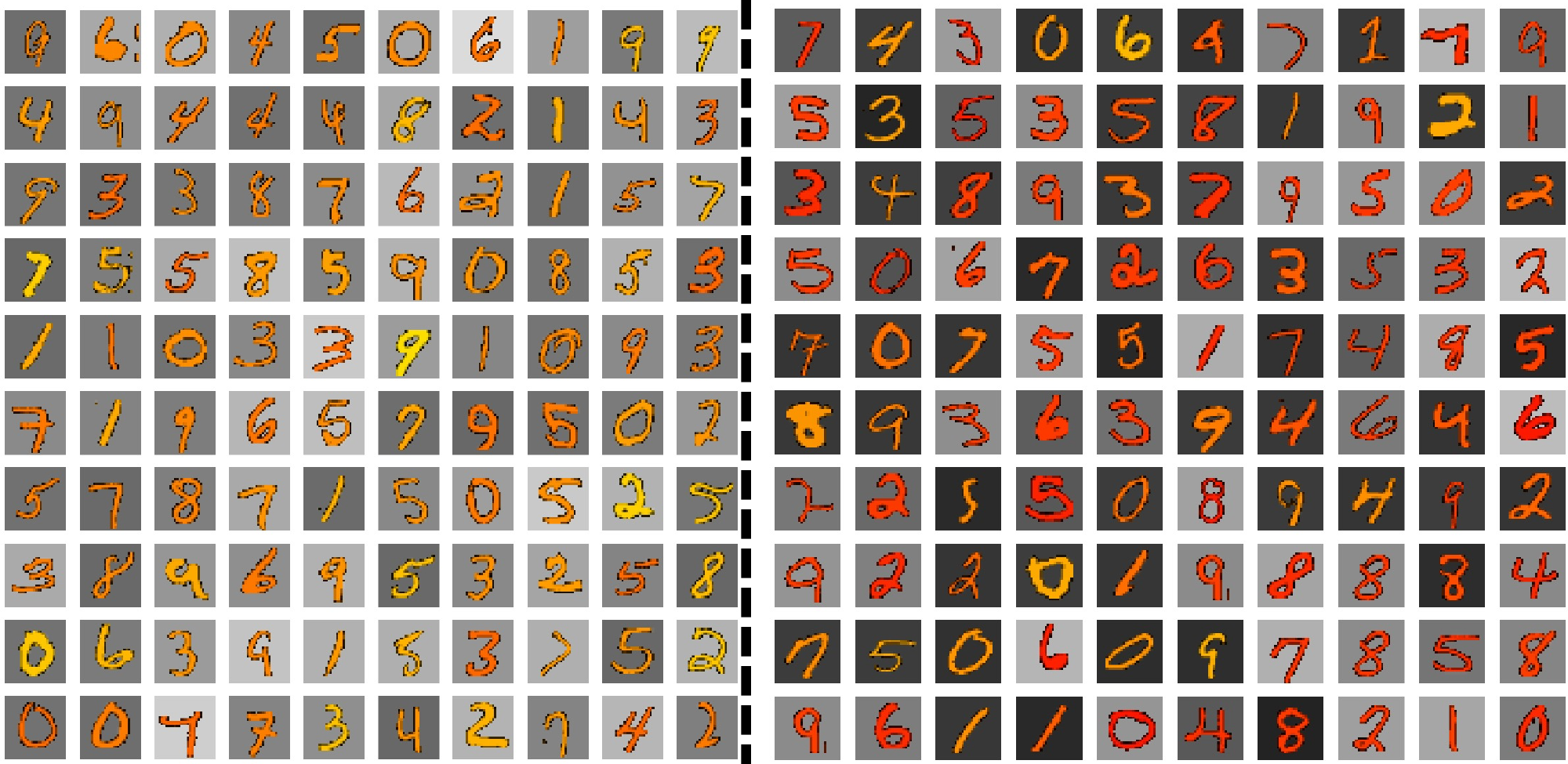}
    \caption{The left figure shows samples from the source environment which has lighter digits and backgrounds while the right figure shows the target environment which has darker digits and/or darker backgrounds}
    \label{fig:high-dim-environments}
\end{figure}

\paragraph{Data Generation}
The base data is the 60,000 grayscale handwritten digits from the MNIST dataset \cite{deng2012mnist}.
We first colored each digit by copying itself along the red and green channel axes with an empty blue channel, yielding an initial dataset of yellow digits.
We then randomly sampled 60,000 points from a two-dimensional Beta distribution with shape parameters, $\alpha = \beta = 5$. 
The first dimension of our Beta distribution represented how much of the green channel would be visible per sample meaning low values would result in a red image, while high values would result in a yellow image.
The second dimension of our Beta distribution represented how white vs. black the background of the image would be, where $0 \coloneqq$ black background and $1 \coloneqq$ white background.

Specifically, the data was generated as follows. 
With $\xvec_{raw}$ representing a grayscale digit from the unprocessed MNIST dataset, a mask of representing the background was calculated $\mathbf{m} = \xvec_{raw} \leq 0.1$, where any pixel value below $0.1$ is deemed to be the background (where each pixel value $\in [0, 1]$).
Then, the foreground (\ie digit) color was created $\xvec_{digit-color} = [(1- \mathbf{m}) \cdot \xvec_{raw}, ~ b_1 \cdot (1- \mathbf{m}) \cdot \xvec_{raw}, \mathbf{0}]$, where $\mathbf{0}$ is a zero-valued matrix matching the size of $\xvec_{raw}$ and $b_1 \sim \text{Beta}(\alpha, \beta)$.
The background color was calculated via $\xvec_{back-color} = [b_2 \cdot \mathbf{m} \cdot \xvec_{raw}, ~ b_2 \cdot \mathbf{m} \cdot \xvec_{raw}, ~ b_2 \cdot \mathbf{m} \cdot \xvec_{raw}]$.
Then $\xvec_{colored} = \xvec_{digit-color} + \xvec_{back-color}$, which results in a colorized MNIST digit with a stochastic foreground and background coloring.

The environments were created by setting the source environment to be any images where $b_1 \geq 0.4$ and $b_2 \geq 0.4$, \ie any colorized digits that had over 40\% of the green channel visible \emph{and} a background at least 40\% white, and the target environment is all other images.
Informally, this split can be thought of as three sub-shifts: a shift which is only reddens the digit, a second shift which only a darkens the background, and a third shift which is both a digit reddening and background darkening.
The environments can be seen in \autoref{fig:high-dim-environments}.

\paragraph{Model}
To encode and decode the colored images, we used a semi-supervised VAE (SSVAE) \cite{siddharth2017learning}.
The SSVAE encoder consisted of an initial linear layer with input size of $3\cdot 28 \cdot 28$ and a latent size of $1024$.  This was then multi-headed into a classification linear layer of size $1024$  to $10$, and for each sample with a label, digit label classification was performed on the output of this layer.
The second head of the input layer was sent to a style linear layer of size $1024$ to $50$ which is to represent the style of the digit (and is not used in classification).
The decoder followed a typical VAE decoder approach on a concatenation of the classification and style latent dimensions.
The SSVAE was trained for 200 epochs on a concatenation of both $\sourceD$ and $\targetD$ with 80\% of the labels available per environment, and a batch size of 128 (for training details please see \cite{siddharth2017learning}).
The transport mapping was then found on the saved lower-dimensional embeddings.

\begin{figure}[!ht]
    \centering
    \includegraphics[width=\textwidth]{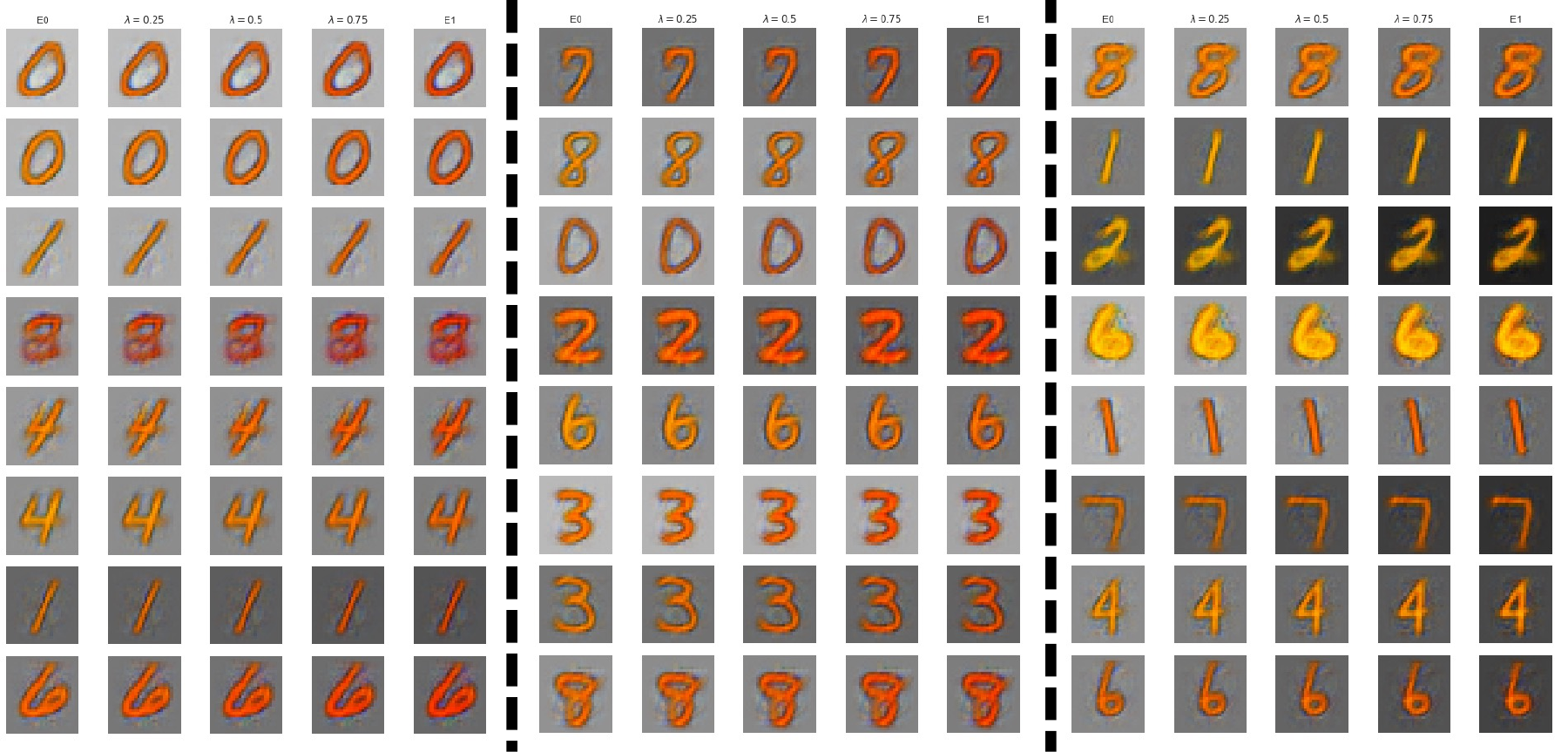}
    \caption{The linear interpolation explanations for the three clusters where the left cluster \emph{seems} to explain the darkening digit shift, the right-most figure explains the shift which darkens the background, and the middle cluster explains the case where both digit and background darkens.
    For each cluster, the left-most digit $\xvec$ is the reconstruction of original encoding from the source distribution, the right-most digit is the cluster-based push-forward of that digit $T(\xvec)$, and the three middle images are reconstructions of a linear interpolations $\lambda \cdot \xvec + (1 - \lambda) \cdot T(\xvec)$ with $\lambda \in \{0.25, 0.5, 0.75\}$.}
    \label{fig:high-dim-cluster-explanations}
\end{figure}

\paragraph{Shift Explanation Results}
Given the shift is divided into three main sub-shifts, we used $k=3$ cluster maps to explain the shift.
We followed the approach given in \autoref{eq:simplified-THIT}, where the three cluster labels and transport were found in the 60 dimensional latent space using the algorithm given in Algorithm 1.
Since our current approach is not able to find a latent space with disentangled and semantically meaningful axes, we cannot use the mean shift information per cluster as the explanation itself (as it is meaningless if the space is uninterpretable).
Instead, we provide an operator with $m$ samples from our source environment and the linear interpolation to the samples' push-forward versions under the target environment, for each cluster.
The goal is for the operator to discern the meaning of each cluster's mean shift by finding the invariances across the $m$ linear interpolations.
The explanations can be seen in \autoref{fig:high-dim-cluster-explanations}.

The linear interpolations from the first cluster (the left of \autoref{fig:high-dim-cluster-explanations}) seem to show a darkening of the source digit, while keeping the background relatively constant.
The third cluster (right-most side of the figure) represents the situation where only the background is darkened but the digit is not.
Finally, the third cluster seems to explain the sub-shift where both the background and the digit are darkened.
However, the changes made in the figures are quite faint, and without \emph{a priori} knowledge of the shift it is possible that this could be an insufficient explanation.
As mentioned in \autoref{sec:appendix-explaining-shifts-in-images}, this could be improved by finding a disentangled latent space with semantically meaningful dimensions, better approximating high dimensional empirical optimal transport maps, jointly finding a representation space and transport map like in \autoref{eq:simplified-THIT}, and more; however, these advancements are out of scope for this work.
We hope that this current preliminary experiment showcases the validity of using transportation maps to explain distribution shifts in images and inspires future work to build upon this foundation.

\end{document}